%% file: main.tex
\documentclass{article}

\usepackage[preprint]{neurips_2025}

\usepackage{hyperref}
\usepackage{url}
\usepackage{xurl}
\usepackage{booktabs}
\usepackage{amsmath,amssymb,amsfonts}
\usepackage{graphicx}
\usepackage{multirow}
\usepackage{makecell}
\usepackage{enumitem}
\usepackage{subcaption}
\usepackage{threeparttable}
\usepackage{xcolor}
\usepackage{colortbl}
\usepackage{microtype}
\usepackage{tabularx}
\usepackage{xspace}
\usepackage{float}
\usepackage{natbib}
\setcitestyle{numbers,square}

\usepackage{fancyhdr}
\AtBeginDocument{%
  \newgeometry{textheight=9in,textwidth=5.5in,top=1in,headheight=42pt,headsep=15pt,footskip=30pt}%
}
\fancypagestyle{empty}{%
  \fancyhf{}%
  \fancyhead[L]{\includegraphics[width=120pt]{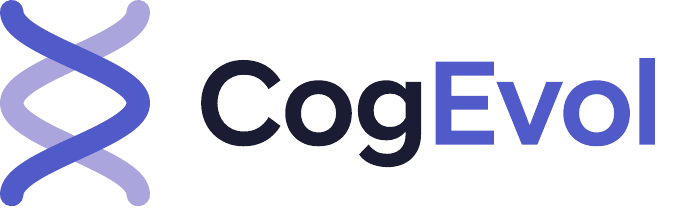}}%
}
\makeatletter
\renewcommand{\@toptitlebar}{\vskip -\parskip}
\makeatother

\newcommand{\model}{CogEvol\xspace}

\title{CogEvol: Towards Efficient and Reliable\\ Learning Environment Generation}

\author{
{CogEvol Team}
\\[3pt]
CogEvol Inc. ~\&~ Tsinghua University
}

\begin{document}

\maketitle

\vspace{-1.5em}
\begin{abstract}
\input{abstract}
\end{abstract}

\input{1_intro}
\input{2_task}
\input{3_sft}
\input{4_rl}
\input{5_evaluation}
\input{6_inference}
\input{7_deployment}
\input{9_conclusion}

\bibliographystyle{abbrv}
\bibliography{ref}

\clearpage
\input{appendix}

\end{document}

%% file: abstract.tex
We present \model, a family of models trained specifically for \emph{Learning Environment Generation}: turning a course brief into a finished learning artifact---structured-JSON slides or self-contained interactive HTML pages---in a single pass. Across 220k production requests, \model completes a slide in a median of 17 seconds and an interactive page in 59, replacing minutes-long multi-turn agent scaffolding. Reliability is enforced rather than hoped for: a production-grounded data pipeline turns real failures into 53,687 verified SFT samples, and a hybrid rule-plus-VLM reward drives GRPO-based RL, hardened after we caught and fixed a reward-hacking episode that produced visually convincing but unplayable games. \model-27B scores 83.7 on slide quality and 63.7 on a 500-case interactive-HTML benchmark with $26.9\times$ fewer parameters than flagship coding models, and, in collaboration with the OpenMAIC team, serves their live production traffic. \model-4B is released openly under the Apache~2.0 license at \url{https://github.com/CogEvol/CogEvol-4B}; external flagships are measured on the same suites under the identical harness. Scaffold editing cuts interactive-page generation cost by a further ${\sim}76\%$, and the full stack runs on domestic Ascend accelerators at application-level parity with A800 GPUs, lowering the unit cost of AI-native education at scale.

%% file: 1_intro.tex
\begin{figure}[!h]
  \centering
  \includegraphics[width=\linewidth]{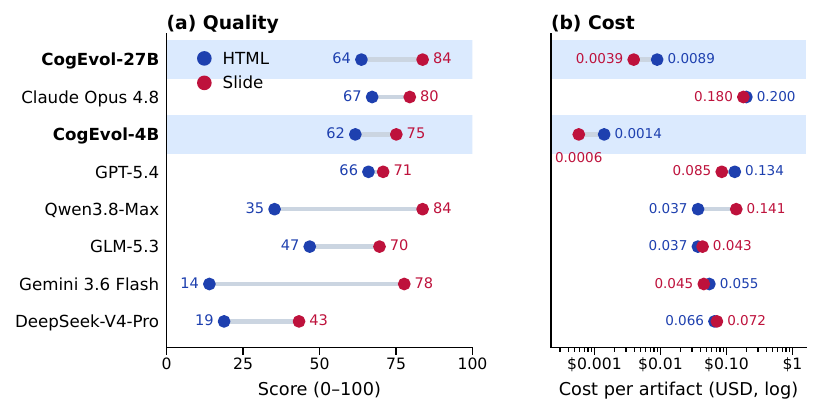}
  \caption{Results of \model-27B, \model-4B, Claude Opus 4.8, GPT-5.4, Qwen3.8-Max, GLM-5.3, Gemini 3.6 Flash, and DeepSeek-V4-Pro on our two suites: (a)~quality on HTML-500 (blue) and slide-std (red), each on a 0--100 scale; (b)~mean API cost per artifact at public list prices, computed from token usage measured on the identical benchmark runs.}
  \label{fig:flagship}
\end{figure}

\section{Introduction}
\label{sec:intro}

\begin{figure}[!t]
  \centering
  \includegraphics[width=\linewidth]{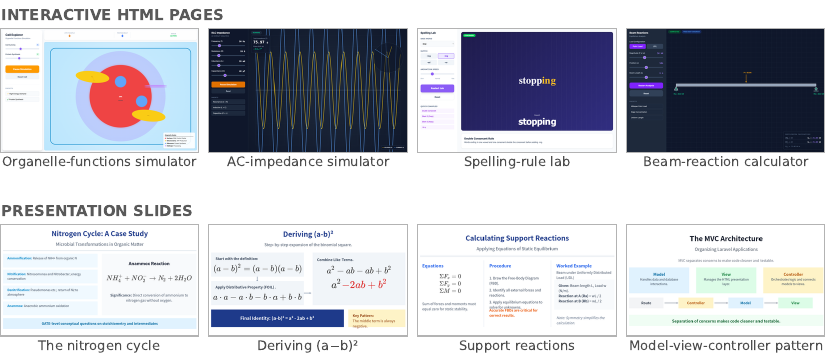}
  \caption{\model output at a glance, from live production traffic on OpenMAIC: eight artifacts generated in a single pass by \model-27B from natural-language course briefs---no agent scaffolding, no human editing. Top: interactive HTML pages---an organelle-functions cell simulator, an AC-impedance circuit simulator (running), a spelling-rule lab, and a beam-reaction calculator. Bottom: slides from generated decks---the nitrogen cycle, deriving the binomial square, support reactions in static equilibrium, and the model--view-controller architecture. Chinese-language examples from the same traffic are in Figure~\ref{fig:showcase-zh} (Appendix~\ref{app:zhshowcase}).}
  \label{fig:showcase}
\end{figure}

\model, short for \emph{Cognitive Co-Evolution}, is a family of models built for education. The name states our long-term goal: humans and machines improving together---models serve learners at scale, and what deployment teaches us feeds back into better models. Education is where we start, not where we stop. Within it, we work on the \emph{learning environment}: the materials that surround a lesson. These materials are shifting from static text to interactive artifacts---slides described in structured JSON, and runnable HTML courseware such as simulations, diagrams, games, and code playgrounds that students can directly manipulate (Figure~\ref{fig:showcase}). Classroom studies have documented the learning benefits of such interactive materials for decades~\cite{richards1992computer,fan2013enhancing,hoon2010effect}. Three gaps keep general-purpose models out of production for this workload: \emph{speed}, \emph{reliability}, and \emph{cost}.

\paragraph{Efficiency.}
General-purpose coding agents can in principle produce such artifacts today: a strong LLM paired with a Claude-Code-style scaffold~\cite{wang2025openhands} will iterate its way to a passable slide or page over many turns of tool use. But the recipe is slow: when a change requires regenerating an entire file, existing systems routinely take 200--600 seconds per edit~\cite{fakhoury2024llm,huang2024new}, and in our own three-task measurement a direct-API editing loop on the same backbone model averaged 152 seconds per edit. Our stack, a purpose-trained model family plus the OpenMAIC harness, completes a full slide in a median of 17 seconds and a full interactive page in a median of 59 seconds, each in a single pass with no agent scaffolding. These are production numbers, not laboratory ones. Over a recent seven-day window of live traffic (August 2026), the production models behind these medians---same architecture and parameter count as \model-27B, and therefore the same serving speed---completed 180k slide generations at a median (P95) of 17s (26s) and 40k interactive pages at 59s (107s).\footnote{Latency statistics from the production serving database (successful calls only, seven-day window); slide pages emit structured JSON (${\sim}1.8$k output tokens on average), interactive pages emit complete HTML documents (${\sim}9$k).} Single-pass generation is not just a convenience: it is what makes interactive courseware usable inside a live class. Iteration stays fast as well: the MAIC-UI editing layer~\cite{tu2026maicuimakinginteractivecourseware} in our harness reduces per-edit latency by $23\times$ (151.7s to 6.3s) compared with direct API calls under the same backbone model.

\paragraph{Reliability.}
Raw coding ability does not transfer to this task. Even strong coding models---GLM-5~\cite{zeng2026glm}, DeepSeek~\cite{deepseekai2025deepseekv32}---fail in ways that surface only at deployment. For slides, models that have never seen the renderer's conventions emit scene graphs that violate its element schema: under a strict validator, none of the external flagships we test renders a single slide, and even with the production pipeline's normalization the best of them lands sixteen points below our SFT-only baseline (Appendix~\ref{app:external}). For interactive pages, one-pass outputs often look impressive---rich visuals, densely stacked features---but break on first contact: dead buttons, unresponsive canvases, simulations that violate basic real-world rules, and content that drifts beyond educational scope. Appearance is easy to fake; dependable interactivity is not. This observation drives our central design principle: \emph{interactivity must be measured, not judged} (Section~\ref{sec:rl}).

\paragraph{Cost and equity.}
Finally, cost decides who gets to use the technology. Simulation-based courseware has always been resource-intensive to author and deploy~\cite{savoldelli2005barriers}; AI generation promised to remove that barrier, but the strongest coding models are enormous---GLM-5 weighs 744B parameters~\cite{zeng2026glm}---and their serving economics price out exactly the users who need educational tooling most. \model is designed around the opposite goal: \model-27B, at 27.7B parameters, is $26.9\times$ smaller than GLM-5 in total parameters,\footnote{744B total vs.\ 27.7B; per-token compute is likewise lower (40B active for GLM-5 vs.\ dense 27.7B).} yet delivers production-grade quality on this task, served as a low-cost API for AI+education developers; \model-4B is open-weight and small enough for on-device deployment. We further adapt the full stack---quantization, operators, scheduling---to domestic accelerators (Section~\ref{sec:deployment}). Low-cost APIs plus deployable open weights lower the unit cost of learning-environment generation enough to reach remote and under-resourced users: teachers in mountain regions, low-income families investing in their children's education, and olympiad training in less-developed areas. Efficient, reliable, and cheap generation is what turns AI education from a premium product into public infrastructure.

\paragraph{The task.}
Education research has long spoken of the \emph{learning environment}---the physical, cultural, and digital setting within which teaching and learning occur~\cite{oecd2013ile}. As instruction moves onto screens, that environment is increasingly made of software, and generative AI is moving into education with it~\cite{wang2025large,alasadi2023generative}. We formalize its automatic construction as \emph{Learning Environment Generation} (LEG; Section~\ref{sec:task}): given a course brief, produce in a single pass either a presentation slide as a renderer-valid structured scene graph, or a self-contained executable interactive HTML page spanning six educational sub-types. Outputs are judged on fidelity, layout, interactivity, and correctness. One request in, one finished learning artifact out; both modalities served by one model behind one interface.

\paragraph{Results.}
On efficiency, \model generates a slide in a median of 17 seconds and a complete interactive page in 59 (production medians over 220k requests), where agent-based pipelines spend minutes per artifact; scaffold editing then cuts interactive-page regeneration cost by a further ${\sim}76\%$, and the MAIC-UI harness accelerates iterative edits by $23\times$. On reliability, \model-27B reaches 63.7 on HTML-500 and 83.7 on slide-std---the best slide score in the \model lineage, 29 points ahead of the best zero-shot flagship slide score. Handed the full 34\,KB design specification, every flagship stays at or below 79.5 except Qwen3.8-Max, which exactly ties 83.7---and then collapses on HTML at 35.3, with 204 of 500 pages dead (Appendix~\ref{app:external}). \model-27B posts zero interactive hard failures on all 500 pages, where Claude Opus 4.8---which edges the HTML overall at 67.2---fails 19 outright. The hardened reward also eliminated the failure mode that motivated it: an earlier reward-hacked checkpoint scores 18.8 on games under the probe, the released model 57.6; human testers saw unusable pages halve (25\% $\rightarrow$ 10\%) and entry failures vanish (2/24 $\rightarrow$ 0/30) after the fix. On cost, Figure~\ref{fig:flagship} puts the two together: at public list prices, \model-27B delivers near-flagship quality at 15--22$\times$ lower per-artifact API cost than Claude Opus 4.8 or GPT-5.4, and \model-4B at ${\sim}100\times$ lower. The full serving stack also runs on domestic Ascend accelerators at application-level parity with A800 GPUs.

\paragraph{Methods.}
The recipe is post-training only: three stages on top of public base models. \emph{Mix SFT} teaches the two output contracts on 53,687 verified conversations distilled from production---teacher-generated scene graphs accepted only after render-and-judge verification, and regenerated pages for the requests the incumbent model failed, accepted only after re-execution (Section~\ref{sec:sft}). \emph{Slide RL} then teaches composition under a hybrid rule-plus-VLM reward, and \emph{interactive-HTML RL} teaches dependable interactivity under a probe-hardened reward---the stage where we caught, and fixed, our reward-hacking episode (Section~\ref{sec:rl}). Both released models are then made cheap to serve: scaffold editing reuses the existing page as scaffolding instead of regenerating from scratch (Section~\ref{sec:inference}), and a deployment-time adaptation layer ports the hybrid architecture to domestic accelerators (Section~\ref{sec:deployment}).

In summary, this report makes the following contributions:
\begin{itemize}[leftmargin=1.5em,itemsep=2pt]
  \item \textbf{Task and benchmarks.} We formalize Learning Environment Generation and build evaluation suites for both modalities: slide-std/slide-short (120+120 topics) and a 500-case interactive-HTML benchmark with executable interaction probes.
  \item \textbf{Efficiency.} Purpose-trained single-turn generation (median 17s per slide, 59s per interactive page across 220k production requests) replaces minutes-long multi-turn agent scaffolding, scaffold editing cuts interactive-page generation cost by a further ${\sim}76\%$ in tokens and latency, and the MAIC-UI harness layer accelerates iterative edits by $23\times$ (Section~\ref{sec:inference}).
  \item \textbf{Reliability.} A hybrid rule-engine + VLM reward system, hardened after a reward-hacking episode on interactive games that we found and fixed, and a one-big-round multi-task recipe that trains both modalities jointly with zero forgetting tax (Section~\ref{sec:rl}).
  \item \textbf{Cost and open release.} \model-27B matches production needs at $26.9\times$ fewer parameters than coding flagships; \model-4B is released openly; the full stack runs on domestic Ascend accelerators with application-level parity (Sections~\ref{sec:deployment}).
\end{itemize}

\paragraph{Open release and evaluation.}
We release \model-4B under the Apache~2.0 license: full-precision weights at \url{https://huggingface.co/CogEvol/CogEvol-4B}, a Q4\_K\_M GGUF build for on-device use at \url{https://huggingface.co/CogEvol/CogEvol-4B-Q4_K_M-GGUF}, and the deployment guide with the MAIC-UI editing harness at \url{https://github.com/CogEvol/CogEvol-4B}. \model-27B is served through a low-cost production API on our MaaS platform; external testing is granted upon request via \href{mailto:contact@cogevol.com}{contact@cogevol.com}. On the application side, OpenMAIC is, to date, the only open-source application whose harness matches the full capability profile of \model. The CogEvol and OpenMAIC teams are separate groups that work closely together: testing, on-device adaptation, and production deployment were all joint efforts. \model is not built exclusively for OpenMAIC---we welcome other AI+Education teams to open their harnesses to us, and we will adapt \model to them as we did for OpenMAIC. The evaluation suites and judge prompts are maintained internally to keep the scoring fixed and the topics uncontaminated; external models are still evaluated on both suites by API submission.

%% file: 2_task.tex
\section{The Learning Environment Generation Task}
\label{sec:task}

\subsection{Task Definition}

\emph{Learning Environment Generation} (LEG) asks a model to turn a course topic into a complete, ready-to-use learning artifact in a single pass. The input is a brief, in either of the two forms production traffic actually takes: a short three-segment description (topic, audience, intent), or a detailed long-form specification that fixes layout regions, per-region content, and visual style. The output is one of two artifacts, each governed by a strict contract.

\paragraph{Modality A: slides as structured JSON.}
A single 16:9 slide is emitted as a JSON scene graph on a $1000\times562$ canvas: eight element types (text, shape, line, image, table, chart, \LaTeX, video), each carrying numeric geometry fields, plus a background. The schema is a \emph{rendering contract}, not a suggestion---the production renderer hard-fails on unrecognized keys, string-typed coordinates, percentages, or invented element types. Charts must use one of nine fixed chart types with an exact data shape; tables follow a cell-level schema with spans; lines carry start/end points instead of bounding boxes.

\paragraph{Modality B: interactive pages as executable HTML.}
The model emits a self-contained HTML document---no external assets, no build step---spanning six educational sub-types: simulations of physical systems, diagrams, games, code playgrounds, 3D visualizations, and structured learning pages. A page succeeds only if it runs: interactions respond, state stays consistent, and the content obeys the physics or mathematics it depicts.

One model serves both modalities; the system prompt carries the contract and selects the modality---slide requests carry the scene-graph schema, interactive requests carry per-sub-type templates. No agent loop mediates between the model and the artifact: one call in, one artifact out.

\paragraph{The name, and its scope.}
The term \emph{learning environment} has a long history in education research---from the design of physical classrooms and learning spaces to the cultures and conditions of a course, and more recently to the digital platforms that host one~\cite{oecd2013ile}. LEG applies the term to the \emph{artifacts} themselves: the task generates the environment's content---its slides and interactive pages---not the platform plumbing (accounts, analytics, distribution) that surrounds them. To our knowledge, generation of such environments has not been formalized as a task before; the related-work boundaries are drawn in the ``Why a new task'' paragraph below.

\paragraph{Quality dimensions.}
We score LEG outputs along four dimensions: \emph{fidelity} (the artifact teaches what the brief asked), \emph{layout} (typography, occlusion, canvas use), \emph{interactivity} (controls act, games are playable), and \emph{correctness} (behavior matches the real-world system being taught, within educational scope). The first two are judged on rendered output; the third is \emph{measured} by executing the artifact and probing its event stream (Section~\ref{sec:eval})---a distinction that shapes both our reward design and our benchmarks.

\paragraph{Why a new task.}
LLM systems in education have so far targeted conversational tutoring~\cite{tu2023littlemu}, lecture-script generation from existing materials~\cite{wang2025educraft}, and interactive learning narratives~\cite{cheng2025oak}---not the generation of executable artifacts under a rendering contract. LEG borders several existing generation tasks without being covered by any of them. Static content generation (documents, images, slide text) requires no executability or interaction; code- and web-UI benchmarks carry no educational semantics and impose no rendering contract; prior slide-generation work evaluates textual outlines rather than renderer-valid scenes. The contract is binding for general-purpose models: the bare Qwen3.8-27B base emits JSON that parses for 118/120 slide briefs yet renders \textbf{0/120} under the strict schema---it invents its own element-field names, a learned convention no model can deduce. Routed through the serving stack's schema normalization, the same outputs do render (118/120)---and reveal what post-training is actually for: content fidelity lands near the family's best (91.7 on the 0--100 scale) while composition collapses (layout 41.5, versus 77.5 after slide RL; Section~\ref{sec:eval} quantifies the same two-layer gap across external flagships). This motivates the dedicated post-training, rewards, and benchmark suite that make up this report.

\subsection{The \model Family}

\model ships in two sizes trained with the same recipe family: \model-4B, the open release, and \model-27B, the production model that, in collaboration with the OpenMAIC team, has served their production traffic since 2026-08. Table~\ref{tab:family} summarizes both; intermediate checkpoints named there are ablation references in Sections~\ref{sec:rl} and~\ref{sec:eval}, not separate products.

\begin{table}[!ht]
  \centering
  \caption{The \model family. Intermediate checkpoints referenced in ablations: \model-4B-SFT, \model-4B-SlideRL, \model-4B-MixRL (mixed-task RL), \model-27B-SFT, \model-27B-SlideRL; Section~\ref{sec:eval} also reports an old-reward variant \model-27B-RL-v1.}
  \label{tab:family}
  \small
  \setlength{\tabcolsep}{4pt}
  \begin{tabularx}{\textwidth}{@{\hspace{5pt}}l X X@{\hspace{5pt}}}
    \toprule
     & \textbf{\model-4B} & \textbf{\model-27B} \\
    \midrule
    Base model & Qwen3.5-4B (dense) & Qwen3.8-27B (hybrid: 48 GDN + 16 full-attn layers, MTP) \\
    Role & open release: weights (Apache~2.0),\newline MAIC-UI editing harness & production serving since 2026-08\newline(low-cost API on our MaaS platform) \\
    \rowcolor{gray!12}
    Post-training & \multicolumn{2}{c}{mix SFT $\rightarrow$ slide RL $\rightarrow$ interactive-HTML RL} \\
    \bottomrule
  \end{tabularx}
\end{table}

\paragraph{Base selection.}
\model-27B starts from Qwen3.8-27B~\cite{qwen3}, a hybrid architecture interleaving 48 gated-delta-net linear-attention layers with 16 full-attention layers and a multi-token-prediction head~\cite{gloeckle2024mtp}; \model-4B starts from the dense Qwen3.5-4B. The 27B base was chosen empirically. Under an identical SFT recipe (same mix-0812 data, same 13,421 updates), Qwen3.8-27B reaches 79.5 on slide-std against 67.7 for Qwen3.6-27B, with slide-contract parse rates of 99.2\% versus 85.8\%; the best Qwen3.6 checkpoint anywhere in its lineage---a different data mix plus its own slide-RL stage---tops out at 81.4, still below the 84.8 the same slide-RL recipe reaches from the Qwen3.8 SFT. Base capability, not update count, sets the ceiling---and the advantage compounds downstream: the slide RL stage on the Qwen3.8 base later transfers $+10.4$pp to interactive HTML without any HTML RL (Section~\ref{sec:rl}). The hybrid architecture raises serving questions of its own, which Section~\ref{sec:deployment} addresses on domestic accelerators.

\paragraph{No pre-training.}
\model is built entirely through post-training. We start from publicly available base models and contribute everything above them: the data pipelines, the hybrid reward system and RL recipes, the evaluation suites and their judge protocols, and the serving infrastructure.

%% file: 3_sft.tex
\section{Post-Training I: Supervised Fine-Tuning}
\label{sec:sft}

\begin{figure*}[!t]
  \centering
  \includegraphics[width=0.98\textwidth]{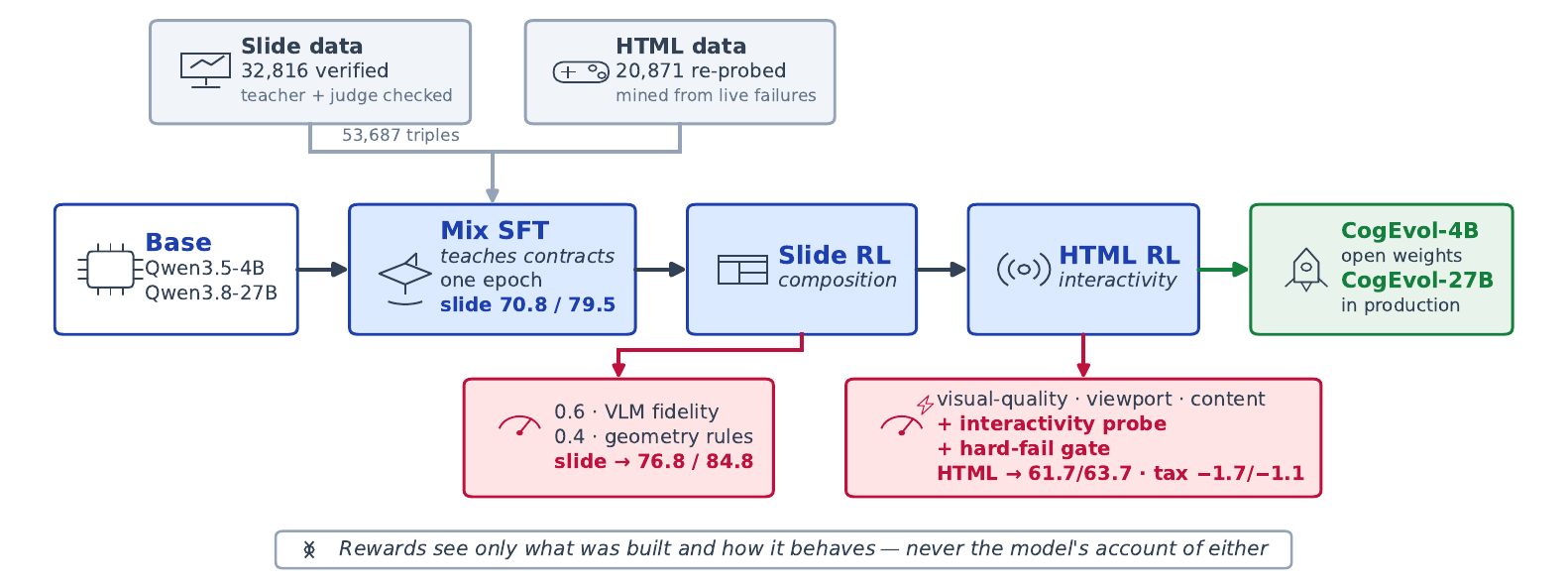}
  \caption{The \model training pipeline. Two execution-aware data pipelines (left) build verified supervision---slides passing render and judge checks, HTML pages surviving a Chromium interactivity probe. Stage~1 teaches the two output \emph{contracts} in one epoch of joint SFT. Stage~2 runs slide RL with a hybrid reward ($0.6$ VLM fidelity on rendered pixels $+$ $0.4$ geometric rules). Stage~3 runs interactive-HTML RL under the hardened reward, whose interactivity probe and hard-fail gate made the difference in Section~\ref{sec:htmlrl}. Scores show slide-std / HTML-500 (0--100) after each stage, as \model-4B / \model-27B.}
  \label{fig:pipeline}
\end{figure*}

\subsection{Production-Grounded Data Pipelines}

The SFT targets of LEG are executable artifacts, not free-form text: syntactic validity is necessary but insufficient---a schema-valid slide can still clip text or overlap tables, and a valid HTML document can still crash on load or expose dead controls. Our two data pipelines share three principles: training prompts derive from real usage, teacher outputs are accepted only after \emph{execution-aware} verification, and the supervision format matches the deployment-time contract.

\paragraph{Slide pipeline: production-seeded synthesis.}
Rather than synthesizing a broad prompt distribution, we start from completed production slide scenes joined with their outlines and image assets. A specification model converts each seed into a self-contained design brief (content hierarchy, layout intent, style, negative constraints, asset placement); data expansion happens at this specification layer, with each round of variants targeting failure modes observed in the previous checkpoint---table and footer collisions, English wrapping, chart-label overlap, formula-heavy and code layouts, dense comparison cards. The final round seeds 500 production scenes with six structural variants each, yielding 3,000 hard layout prompts; fixed benchmark topics are excluded before teacher generation to prevent train--test contamination. Teacher candidates (Gemini 3.1 Pro and 3.5 Flash~\cite{team2023gemini}) emit the final scene graph directly; each candidate passes JSON parsing, schema validation, a conservative canonicalization step, a production-render pass, and a multimodal judge scoring fidelity and layout on a five-point scale. Canonicalization matters more than it sounds: in a control experiment it recovered 26 of 28 wrongly rejected candidates and lifted schema-valid, renderable slides from 92 to 118 of 120---format strictness is a poor proxy for visual quality. A best-of-two arbitration across the two teachers selects one target per brief; the 2,973 selected targets average 4.185 fidelity / 4.280 layout, versus 4.179/4.010 for the stronger single route. After deduplication, tripling the 1,989 hardest examples, and an 8,192-token limit, the slide corpus holds \textbf{32,816} rows (median 2,472 tokens).

\paragraph{HTML pipeline: production failure mining.}
The interactive-HTML corpus concentrates supervision where the incumbent production model actually fails. We exported 119,122 \texttt{interactive-self} generations and executed each in an isolated Chromium probe; 117,309 completed, and 25,475 (21.7\%) exhibited hard failures---initialization crashes, missing interaction contracts, or fully unresponsive controls. Gemini 3 Flash regenerated a complete page for each of the 24,937 resolvable failing requests, and every regeneration was re-executed under the same probe: 17,561 passed directly (72.8\%). Auditing the rejects exposed a systematic false positive in the original contract check: it recognized controls only through slider-style identifiers, though valid simulations also use selects, checkboxes, and semantic handlers. A conservative recovery pass re-admitted 4,412 examples. The usable corpus is \textbf{21,973}; a 16,384-token limit (full pages are long: median 8,988 tokens) trims it to \textbf{20,871}, and a deterministic \texttt{postMessage} bridge listener is inserted into non-simulation targets to align the training representation with the runtime interface.

\paragraph{Known biases.}
Execution-aware filtering improves supervision but shapes it. The slide targets skew safe and sparse (mean 14.4 elements; 68.5\% at most 16), and the specification model's natural, English-rich briefs differ from the production brief writer---a train--serve mismatch that blunted initial deployment transfer. The HTML corpus is 69.8\% simulations, 2.4\% code tasks, and contains no 3D examples. These biases are exactly why SFT alone cannot carry HTML quality (next subsection), and why the RL stages of Section~\ref{sec:rl} operate on prompt distributions rather than fixed datasets.

The final SFT mixture is \textbf{53,687} conversations: 32,816 slides + 20,871 interactive pages, each a (system contract, user brief, verified artifact) triple.

\subsection{Mixture and Base-Model Ablations}

The mixture itself went through three revisions. The first (73,687 rows, HTML-heavy and without system prompts) caused outright format confusion---the model could not tell which modality a request wanted, and slide contract-compliance fell to 61.7\%. Curating the HTML share to 53,687 and shuffling restored compliance to 83.3\%; repairing the system prompts (each HTML sub-type now carries its production template) brought it to 97.5\% on 4B while slide quality simultaneously recovered to 70.8, within two points of the slide-only specialist baseline (72.8)---joint training, done carefully, costs little in either modality.

Table~\ref{tab:sft-ablation} summarizes the runs behind the two released SFT checkpoints. Two observations matter for everything that follows. \textbf{(1) The mix recipe suppresses HTML with optimizer updates.} Under the pre-hardening reward of that period (comparisons within the period only), HTML quality declines monotonically with update count on Qwen3.6-27B---88.4 at 6,710 updates, 83.7 at 10,711, ${\sim}73.5$ at 13,421---whether the extra updates come from more passes or a smaller batch; the Qwen3.8 run lands at the same 73.2 regardless of base, sitting \emph{below} its own bare base (78.3), having ceded ground precisely where the base was strongest (code $-7.0$pp, game $-17.9$pp). SFT reliably teaches the two output contracts but cannot raise interactive quality; the HTML ceiling is left to RL. \textbf{(2) Base capability dominates.} Under the identical recipe (same mix-0812 data, same single-node 13,421-step schedule), Qwen3.8-27B reaches 79.5 on slide-std against 67.7 for Qwen3.6-27B, with contract parse rates of 99.2\% versus 85.8\%; the best Qwen3.6 checkpoint anywhere in its lineage (a different data mix plus its own slide-RL stage) tops out at 81.4, still below the 84.8 the same slide-RL recipe reaches from the Qwen3.8 SFT.

\begin{table}[!ht]
  \centering
  \caption{SFT ablations. Slide-std (0--100; same Gemini-judge protocol for all rows) and contract parse rate. HTML-500 overall under the hardened reward, 0--100, directly comparable with Table~\ref{tab:main}; dashes: not yet scored. The pre-hardening HTML decline across these runs is reported in the text.}
  \label{tab:sft-ablation}
  \small
  \setlength{\tabcolsep}{4pt}
  \begin{tabular}{@{\hspace{5pt}}llcccc@{\hspace{5pt}}}
    \toprule
    Base & Schedule & Slide & Parse & HTML & Note \\
    \midrule
    Qwen3.5-4B & GBS8, 6,710 steps (1 ep) & \textbf{70.8} & 97.5\% & \textbf{52.8} & \textbf{\model-4B-SFT} \\
    \midrule
    Qwen3.6-27B & GBS8, 6,710 steps (1 ep) & 70.2 & 80.8\% & --- & \\
    Qwen3.6-27B & GBS8, 10,711 steps (1.6 ep) & 72.2 & 95.8\% & --- & resume $+4$k \\
    Qwen3.6-27B & GBS4, 13,421 steps (1 ep) & 67.7 & 85.8\% & --- & \\
    Qwen3.8-27B & GBS4, 13,421 steps (1 ep) & \textbf{79.5} & 99.2\% & \textbf{61.2} & \textbf{\model-27B-SFT} \\
    Qwen3.8-27B & bare (no SFT) & 66.6 & 98.3\% & --- & strict 0/120; layout 41.5 \\
    \bottomrule
  \end{tabular}
\end{table}

Why, then, does \model-27B-SFT ship from the GBS4 schedule that scored lowest for Qwen3.6-27B? Because that arm was never a recipe candidate---it is a control. It reproduces the incumbent production schedule (GBS4, single node, 13,421 steps) on mix-0812 data, and its failure to rescue Qwen3.6-27B is exactly what falsified the hypothesis that update count, rather than data or base, drove earlier results. The Qwen3.8-27B run then deliberately reused the identical schedule so that the base is the only variable: the 79.5-vs-67.7 slide gap and the 99.2\%-vs-85.8\% parse gap are attributable to the base swap alone, and the resulting checkpoint set the best SFT slide score to date. On Qwen3.6 the dual-node GBS8 schedule does beat GBS4 on both modalities, but no Qwen3.8 GBS8 arm was trained, so that ordering is established on one base only. No schedule variation reversed the HTML decline---which settles the recipe's shape: SFT stays a single contract-teaching pass, and the pursuit of quality moves to RL on top.

%% file: 4_rl.tex
\section{Post-Training II: Reinforcement Learning}
\label{sec:rl}


All RL stages share one setup: GRPO~\cite{shao2024deepseekmath,yu2025dapo} on the slime framework~\cite{slime}, 8--16 H800 GPUs, eight prompts $\times$ eight samples per rollout batch (group size 8), KL coefficient $10^{-3}$ against the stage's initial policy, learning rate $10^{-6}$, thinking disabled, 250 rollouts per stage. Because a candidate can be scored only after it exists visually---slides rendered to PNG, pages loaded in Chromium, probed, and judged from screenshots and probe traces---reward computation dominates wall-clock cost.

\subsection{A Hybrid Reward System for Structured Visual Generation}
\label{sec:rewarddesign}

One design decision precedes every other: \textbf{deterministic dimensions are scored by rules, subjective dimensions by a vision-language judge, and no judge takes the model's word for anything.} The visual judges receive only rendered pixels and probe traces. Where the HTML reward's content judge does consult the page source, it reads a style-stripped listing as structural evidence, checking whether the control a prompt asked for was actually built, not the model's account of having built it. Rewards are positive scores to be earned, not penalties to be deducted, and every scale is anchored to concrete grade descriptions to keep the judge stable across runs.

\paragraph{Slide reward.}
$R_{\mathrm{slide}} = 0.6 \cdot \mathrm{VLM} + 0.4 \cdot \mathrm{rule}$, both terms on a 0--5 scale. The rule engine encodes geometric ground truth: canvas utilization, element collision, table overflow and sibling occlusion, chart geometry. It evolved through five versions, each fixing what the policy had just learned to exploit---most notably a \emph{pseudo-chart penalty} added when the model began faking charts with styled text to dodge chart rules, and a coordinate-normalization-order bug whose repair finally gave the rule term enough signal to constrain layout. The VLM judge scores content fidelity from the rendered slide; a small $\pm0.03$ advantage jitter prevents identical judge scores within a GRPO group from zeroing the gradient.

\paragraph{Interactive-HTML reward.}
An interactive page can fail in ways that do not overlap, so the reward scores each failure mode separately and weights it by what it costs a student. Two vision-language judges read the rendered desktop screenshot: a \emph{visual quality} judge (0.4) rates layout, readability, and aesthetics, and a \emph{content} judge (0.3) rates instruction fidelity, pedagogical value, and scientific correctness, each on an anchored five-point scale. A \emph{viewport} pair (0.1 each) sets the desktop rendering against tablet and mobile renderings and flags content that vanishes, collides, or scales wrongly at narrow widths. An \emph{interactivity} term (0.3) reports not a judgment but a measurement: what a Playwright-driven Chromium instance observed when it operated the page's controls itself. Each term is mapped to a defect score in $[0,1]$ and the reward is one minus their weighted mean; the probe instrumentation and hard-fail gate behind the interactivity term are detailed in Appendix~\ref{app:probe}.

\paragraph{Scoring discipline.}
Reward versions define the score scale, so numbers scored under different versions are never compared: whenever the reward changes, every baseline is re-scored under the new version, and all HTML scores in this paper (Tables~\ref{tab:sft-ablation},~\ref{tab:main}) come from the hardened version. The same caution applies to the evaluation harness: an early harness bug sent the system template instead of the user topic, silently inflating RL gains, and we now verify instruction-following structurally---all models produce 500/500 unique titles---rather than assume it.

\subsection{Slide RL}

Slide RL taught us three things that shaped everything downstream.

\textbf{(1) Fidelity and layout trade off until the rule signal is trustworthy.} Early rounds bought fidelity at the cost of layout: the VLM holds 60\% of the weight, and with the rule engine weakened by its normalization bug the policy optimized content at the expense of composition (fidelity $+0.46$, layout $-0.31$ at the worst point; slides with severely broken layout rose to 45\%). Only after the rule repair did the two dimensions co-move, and a 4B checkpoint finally beat its SFT start on both simultaneously.

\textbf{(2) Brief quality sets the fidelity ceiling.} Training on topic names alone produces a judge signal too noisy to learn from---the VLM cannot discriminate ``theme vs.\ content'' finely, and gains stall. Switching the prompt pool to model-written detailed briefs (with benchmark IDs excluded) unlocked the fidelity lifts; the richest brief source lifted 4B fidelity to its all-time high.

\textbf{(3) Prompt diversity is a training signal, not decoration.} Purifying the brief pool to a single generator---intended as de-noising---collapsed GRPO: stylistically uniform briefs made the eight samples within a group converge to similar scores, advantages approached zero, and performance fell below the SFT baseline within 200 steps. Filtering short briefs while keeping multiple generators' styles restored health. A corollary emerged at 27B: continuing a converged run on fresh same-distribution data never beat the earlier checkpoint---we always ship the checkpoint at convergence, not after.

The shipped slide stages: on 4B, slide RL lifts the mix-SFT start from 70.8 to \textbf{76.8} on slide-std; on 27B, the same recipe on the Qwen3.8 SFT reaches \textbf{84.8}, with layout 77.5 the highest of the entire lineage. One result from this stage quietly set up the next: 27B slide RL---trained on slide data only, with no HTML in the loop---lifted interactive HTML by $+10.4$pp over its SFT start. The two modalities share representations deeply enough that a single-modality stage transfers.

\subsection{HTML RL and the Game Reward-Hacking Episode}
\label{sec:htmlrl}

Interactive-HTML RL produced the central finding of this report, in five acts.

\textbf{Act 1: strictness first.} The rebuilt judge (Section 4.1) re-based all scores ${\sim}25$pp downward. Under it, two data-driven 4B runs delivered real gains: failure-mined data first, then a weak-type-weighted round whose gains tracked its weighting almost exactly (code $+9.8$pp, 3D $+5.1$pp on the targeted types). By then the eval-harness bug of Section 4.1 had also been found and fixed, cutting the apparent two-run increment from $+11.6$pp to a real $+4.8$pp---the difference was ``template-polishing'' that no-topic evaluation had rewarded.

\textbf{Act 2: a regression that resists more data.} The third run doubled down on the weakest type---games constituted a third of its training batch, the largest share of any type. Games regressed by $-12.1$pp, the worst single-type collapse of the campaign, while every weighted non-game type gained. The control was the first run, whose game-light data had left games unchanged. Data was not the lever; the reward was.

\textbf{Act 3: diagnosis.} Every component of the reward---visual quality, dual-viewport screenshots, content---judged \emph{static renderings}. Nothing in it measured what makes a game: responding to input, enforcing rules, closing its feedback loop. A policy optimizing that reward on game prompts should converge on exactly what we observed: pages whose opening frame screenshots beautifully and whose interaction is broken.

\textbf{Act 4: the fix, as a pure A/B.} We hardened the reward---always-on interactivity probe plus the hard-fail gate, both detailed in Appendix~\ref{app:probe}---and retrained from the same checkpoint, same data, same schedule: the only variable was the reward. Games reversed from $-12.1$pp to $+5.8$pp; 3D rose $+15.9$pp; overall HTML lifted $54.2 \rightarrow 61.7$. The hardening also \emph{protected} the other modality: from the same slide-RL start (76.8), the hardened run pays only a $-1.7$ slide tax while its old-reward twin pays $-4.0$ (72.8)---suppressing visually loud but hollow output styles, it seems, preserves the other modality's aesthetics; at 27B both reward versions pay the same smaller tax instead ($-1.1$; Section~\ref{sec:eval}). The reward's \emph{taste} transfers across tasks.

\textbf{Act 5: the smoking gun.} The 27B old-reward run---same recipe as \model-27B, trained before hardening---had looked merely mediocre on games under the old judge. Re-scored under the hardened reward, its game score is \textbf{18.8}: a $-36$pp collapse the screenshot judge had masked as $-9.7$pp. Its non-game mean matches the hardened run's increment exactly ($+3.2$pp); the old reward learned everything \emph{except} playability, and quietly unlearned that. We disclose this checkpoint in Table~\ref{tab:main} as the cautionary twin.

The principle that survives: \emph{interactivity must be measured, not judged}. The probe is now a permanent component of both the reward and the benchmark gate (Section~\ref{sec:eval}).

\subsection{One-Big-Round Multi-Task RL}

The positive transfer of Section 4.2 and the forgetting tax of Section 4.3 suggest the two modalities want to be trained together. One-big-round RL does exactly that: slide and HTML prompts mixed 50/50 in every batch, one policy, a reward router dispatching each sample to its modality's reward (unroutable requests fail loudly rather than scoring zero). GRPO's group-relative advantage makes the modality-scale mismatch harmless---advantages are computed within same-prompt groups, so the policy never compares a slide score against an HTML score.

Trained on 4B from a pure SFT checkpoint for one epoch, the single round reached slide \textbf{74.1}---within three points of the serial pipeline's post-slide-RL peak (76.8)---while lifting HTML to \textbf{59.0} with games intact (53.6, versus 37.3 for the pre-hardening serial twin). Zero forgetting tax, from an SFT start, in one round; the router ran 500 steps without a miss. The honest ledger: on HTML alone the serial recipe still wins (61.7 vs.\ 59.0), and human inspection of a 100-case side-by-side favored the serial model's pages; per-task gradient is simply halved, and the long-tail types (learning pages, code) pay for it. \model ships the serial recipe; one-big-round stands as the resource-constrained alternative that buys joint quality at no tax, with 27B validation left to future work.

\subsection{Final Recipes: \model-4B and \model-27B}

Both released models follow the same three-stage serial recipe---mix SFT (Section~\ref{sec:sft}) $\rightarrow$ slide RL $\rightarrow$ interactive-HTML RL under the hardened reward---differing only in base and scale:

\begin{itemize}[leftmargin=1.5em,itemsep=2pt]
  \item \textbf{\model-4B}: Qwen3.5-4B $\rightarrow$ mix SFT $\rightarrow$ slide RL $\rightarrow$ HTML RL. HTML-500 \textbf{61.7} (from 52.8 at SFT), slide-std \textbf{75.1}; the forgetting tax of the final stage is $-1.7$.
  \item \textbf{\model-27B}: Qwen3.8-27B $\rightarrow$ mix SFT $\rightarrow$ slide RL (84.8, lineage-best layout) $\rightarrow$ HTML RL. HTML-500 \textbf{63.7}---the best in the \model lineage---games \textbf{57.6}, slide-std \textbf{83.7} after a $-1.1$ serial tax; and its old-reward twin pays exactly the same tax (83.7), pinning that cost on the serial stage itself rather than the reward version.
\end{itemize}

\model-27B, deployed jointly with the OpenMAIC team, has served their production traffic since 2026-08-24, replacing its predecessor behind the same API. The recipe's one-sentence summary: SFT teaches the contracts, slide RL teaches composition, hardened-reward HTML RL teaches interactivity---and what the reward cannot measure, RL will quietly destroy.

%% file: 5_evaluation.tex
\section{Evaluation}
\label{sec:eval}

\subsection{Benchmark Construction}

LEG is a new task, so its evaluation is built rather than borrowed: public benchmarks neither impose our rendering contract nor measure interactivity. Our suite has three components, maintained internally and scored centrally rather than released.

\paragraph{Slide: slide-std and slide-short.}
Two 120-topic sets, one per production distribution. \emph{slide-std} carries detailed long-form briefs (the SFT distribution); \emph{slide-short} carries short three-segment requests (the RL distribution; seed-42 sampled, verified disjoint from training data). Each model generates one slide per topic (temperature 0); every output is rendered by the production renderer---contract violations render nothing and score zero---and the rendered PNG is judged by Gemini 3.1 Pro on fidelity and layout, each on an anchored 0--5 scale. We report each dimension $\times$20 on a 0--100 scale and their mean as the summary (Table~\ref{tab:main}).

\paragraph{Interactive HTML: HTML-500 with a probe gate.}
500 cases spanning the six educational sub-types in production proportions (simulation 197, learning pages 133, diagrams 72, games 60, 3D 20, code playgrounds 18), each carrying its production system prompt, one generation per model (temperature 0.7, 16,384 max tokens, thinking off). Before any judge sees a page, a deterministic probe gate executes it: a Playwright-driven Chromium instance drives the page's interactions and reports initialization crashes, dead controls, and unresponsive games---the same instrumentation as the reward's interactivity probe (Section~\ref{sec:rl}; Appendix~\ref{app:probe}), because interactivity is a measured property, not a judged one. Hard failures score zero regardless of appearance. Surviving pages are scored by the hardened reward's full composite (visual quality, dual-viewport, content, probe), reported $\times$100.

\paragraph{Scoring discipline.}
All HTML scores in this paper are produced by one fixed configuration of the hardened reward---the same scorer used in RL training---so every number is directly comparable across tables. Instruction-following is verified structurally (each model produces 500/500 unique page titles) rather than assumed. Ablations over recipes and reward versions appear alongside the training results (Sections~\ref{sec:sft},~\ref{sec:rl}); this section reports endpoints.

\paragraph{External evaluation.}
The suites and their scoring pipeline are maintained internally rather than released: the HTML scorer is the production RL reward itself, and the topics must stay uncontaminated. External models can still be measured on both suites: developers provide an API endpoint, we run the identical harness end to end, and all external rows in this paper were produced by this service.

\subsection{Main Results}

\paragraph{Setup.}
One generation per model throughout. HTML-500 columns are scored by the single fixed hardened-reward configuration of Section~\ref{sec:eval} (0--100); slide columns are judged on the rendered output, and unrenderable outputs score zero. \model-27B-RL-v1 is the same recipe as \model-27B trained against the pre-hardening reward (Section~\ref{sec:rl}). External flagships run the identical harness and scorer with one difference: their slide columns use the full 34\,KB specification, the strongest fair condition for a model that has not learned the contract (the two prompt settings are defined in Appendix~\ref{app:external}; zero-shot slim-contract results are Table~\ref{tab:external}). The HTML composite is scored by a Qwen3.8-family 27B VLM---the same family as the Qwen3.8-Max baseline, which we disclose here; the Claude endpoints are run at their default temperature.

\begin{table}[!ht]
  \centering
  \caption{Main results on our internal suites and external flagships under the Section~\ref{sec:eval} protocol. Overall: the mean of the HTML Avg and Slide Avg columns. Bold marks the best score per column, underline the second best (ties share a rank).}
  \label{tab:main}
  \small
  \setlength{\tabcolsep}{1.6pt}
  \begin{tabular}{@{\hspace{5pt}}lccccccc|ccc|c@{\hspace{5pt}}}
    \toprule
    \multirow{2}{*}{Model} & \multicolumn{7}{c}{HTML-500 by sub-type} & \multicolumn{3}{c}{Slide (std)} & \multirow{2}{*}{Overall} \\
    \cmidrule(lr){2-8} \cmidrule(lr){9-11}
     & sim & diag & game & code & 3d & learn & Avg & Fid & Lay & Avg & \\
    \midrule
    \model-4B-SFT & 52.7 & 60.6 & 48.0 & 52.3 & 38.0 & 53.2 & 52.8 & 79.7 & 62.0 & 70.8 & 61.8 \\
    \model-4B-SlideRL & 55.4 & 59.5 & 48.0 & 54.7 & 37.4 & 54.7 & 54.2 & 84.3 & 69.3 & 76.8 & 65.5 \\
    \model-4B-MixRL & 63.0 & \underline{66.0} & 53.6 & 50.5 & 57.0 & 53.1 & 59.0 & 80.2 & 68.0 & 74.1 & 66.6 \\
    \model-4B & 64.6 & \textbf{67.4} & 54.5 & 54.3 & 53.3 & 60.1 & 61.7 & 82.8 & 67.3 & 75.1 & 68.4 \\
    \midrule
    \model-27B-SFT & 64.1 & 60.8 & 52.1 & 53.6 & 58.6 & 62.5 & 61.2 & 89.3 & 69.7 & 79.5 & 70.4 \\
    \model-27B-SlideRL & 62.8 & 62.1 & 54.6 & 50.7 & 58.8 & 60.5 & 60.5 & 92.2 & \textbf{77.5} & \textbf{84.8} & 72.7 \\
    \model-27B-RL-v1 (old reward) & 64.7 & \underline{66.0} & 18.8 & 59.1 & 57.6 & 67.2 & 59.6 & 90.8 & 76.5 & \underline{83.7} & 71.7 \\
    \model-27B & 64.4 & 64.4 & \textbf{57.6} & 53.6 & 64.8 & 66.1 & 63.7 & \textbf{93.0} & 74.3 & \underline{83.7} & \textbf{73.7} \\
    \midrule
    GPT-5.4 & \textbf{70.5} & 49.7 & \underline{56.7} & \underline{64.5} & \underline{67.6} & \underline{72.6} & \underline{66.0} & 90.0 & 51.7 & 70.8 & 68.4 \\
    Qwen3.8-Max & 39.4 & 20.1 & 30.5 & 56.0 & 41.4 & 36.0 & 35.3 & \underline{92.7} & 74.7 & \underline{83.7} & 59.5 \\
    DeepSeek-V4-Pro & 11.8 & 26.2 & 27.4 & 49.8 & 4.9 & 21.3 & 18.8 & 45.0 & 41.7 & 43.3 & 31.1 \\
    Claude Opus 4.8 & \underline{70.1} & 65.2 & 48.3 & \textbf{65.8} & \textbf{70.0} & \textbf{72.8} & \textbf{67.2} & 84.7 & 74.3 & 79.5 & \underline{73.3} \\
    Gemini 3.6 Flash & 10.7 & 5.7 & 35.5 & 55.9 & 13.5 & 8.2 & 14.0 & 80.3 & \underline{75.0} & 77.7 & 45.9 \\
    GLM-5.3 & 57.0 & 31.8 & 41.5 & 56.2 & 59.7 & 57.7 & 46.8 & 77.5 & 61.7 & 69.6 & 58.2 \\
    \bottomrule
  \end{tabular}
\end{table}

Four readings of Table~\ref{tab:main}. \textbf{(1) Post-training increments.} On 4B, HTML RL lifts overall reward from 52.8 (SFT) to 61.7 ($+9.0$pp), with the largest gains exactly where the SFT model was weakest---simulation ($+11.9$pp, the dominant sub-type) and 3d ($+15.3$pp); on 27B, the SFT is a stronger start and RL adds $+2.5$pp overall. Slide RL alone does \emph{not} buy HTML capability on 27B (61.2 $\rightarrow$ 60.5, $-0.6$pp), confirming the two modalities need their own RL stages. \textbf{(2) The serial forgetting tax is small, specific to the HTML stage, and lands entirely on layout.} The serial recipe (slide RL $\rightarrow$ HTML RL) costs $-1.1$ slide points on 27B (84.8 $\rightarrow$ 83.7)---fidelity actually rises (92.2 $\rightarrow$ 93.0) while layout falls (77.5 $\rightarrow$ 74.3): the HTML stage polishes content but dilutes composition. The old-reward v1 pays exactly the same tax (83.7), showing the cost belongs to the serial stage itself, not the reward version; the mixed-task recipe holds slide at 74.1 on 4B. \textbf{(3) The game column is the reward-hacking exhibit.} \model-27B-RL-v1 collapses on games (18.8) while scoring \emph{highest} on code (59.1) and learning pages (67.2)---exactly the ``fluent but broken'' signature; the hardened reward restores games to 57.6, the best of any model here, at the cost of some code (Section~\ref{sec:rl} quantifies the trade-off).

\textbf{(4) External flagships.} The external rows run the identical harness, and their slide columns already run under the generous condition---the full 34\,KB design specification our distillation teacher uses (the zero-shot slim-contract setting, sixteen points lower for the best flagship, is Table~\ref{tab:external} in Appendix~\ref{app:external}). Under it, exactly one model reaches \model-27B: Qwen3.8-Max, the flagship of the same family our base belongs to, lands precisely on 83.7 (92.7/74.7)---but only with the 34\,KB document in hand (zero-shot on the slim contract it scores 23.9), whereas \model-27B needs the 310-word training contract alone; every other flagship stays at or below Claude Opus 4.8's 79.5. And the same model collapses on the other modality: 35.3 on HTML-500 with 204 of 500 pages dead at the probe, where \model-27B scores 63.7 with zero hard failures. Claude Opus 4.8 posts the best external HTML average (67.2 vs.\ 63.7, leading four of six sub-types) while failing 19 pages outright on dead interactions, and GPT-5.4 (66.0) fails 13; GLM-5.3 shows the same fluent-but-broken split---its surviving pages score mid-pack, but 115 of 500 are dead (71 at the probe, 44 truncated at the 16{,}384-token budget)---no model leads both modalities under either condition.

\subsection{Human Evaluation}

Automated rewards compress a page into a number; production users experience the whole page. Two rounds of internal manual testing surround the fix: an earlier pre-hardening build (the slide-RL stage plus the old-reward HTML stage---the system before Section~\ref{sec:rl}) and \model-27B itself. Each round generated complete courses on the local OpenMAIC deployment and graded every interactive page by hand. The rounds are directional rather than strictly controlled (prompts differ slightly; $n{=}24$ vs.\ $n{=}30$), and we report them as such.

\begin{table}[!ht]
  \centering
  \caption{Manual testing of interactive pages on the local OpenMAIC deployment. A page is fully usable only if its complete interaction chain works; \emph{unusable} = cannot be entered, blank canvas, or dead core flow. The pre-hardening build pairs the slide-RL stage with the old-reward HTML stage; the other build is \model-27B. Not a strict A/B (prompts differ between rounds, small samples)---read as directional.}
  \label{tab:human}
  \small
  \begin{tabular}{lcc}
    \toprule
     & Pre-hardening ($n{=}24$) & \model-27B ($n{=}30$) \\
    \midrule
    Fully usable & 14 (58.3\%) & 20 (66.7\%) \\
    Core flow usable, minor defects & 4 (16.7\%) & 7 (23.3\%) \\
    Unusable & 6 (25.0\%) & 3 (10.0\%) \\
    \midrule
    Page cannot be entered & 2 & \textbf{0} \\
    Blank main canvas & 4 & 3 \\
    Element stacking / offset & 2 & 4 \\
    Language mixing or garbled text & 6 & 4 \\
    \bottomrule
  \end{tabular}
\end{table}

The headline row is the one Section~\ref{sec:rl} predicts: pages that cannot be entered go from 2/24 to \textbf{0/30}, and all eight games generated in the second round were playable on entry---the interactivity hardening shows up in human hands, not only under the probe. Fully usable pages rise from 58.3\% to 66.7\% while unusable pages halve (25\% $\rightarrow$ 10\%). The defects that persist are as informative: language mixing (6/24 $\rightarrow$ 4/30) and occasional layout stacking survive both rounds---problems the current reward does not price, and therefore did not fix. On slides, the rounds are flat (chart-overlap and theme-drift bad cases recur on the same prompts), consistent with the small slide delta between the two builds (84.8 $\rightarrow$ 83.7 on slide-std).

\subsection{Reward--Human Agreement}
\label{sec:rewardhuman}

The reward of Section~\ref{sec:htmlrl} is both the training signal and the benchmark scorer, so its agreement with human judgment must be verified. We measure how closely reward rankings correspond to independent human ratings, and we used the same measure to choose among candidate judge prompts during development.

\paragraph{Calibrating the judge.}
Without anchored score descriptors, the VLM judge drifts upward on educational HTML pages: with nothing defining what an ordinary or a good page looks like, scores pile up near the top of the scale. A reward with this property gives the policy nothing to learn from---the outputs it cannot discriminate are exactly the ones RL must learn to separate. Judge prompt revisions were therefore evaluated on the score distributions they produced over a held-out set of pages, not on how their wording read.

For two pre-RL baselines of 500 pages each, mean scores and ceiling occupancy (the fraction of pages scoring 4 or 5 out of 5) under both the preceding and current prompts are reported in Table~\ref{tab:reward-dist}. Under the preceding prompts, the Qwen3.8-27B baseline receives a mean of 4.88 on scientific correctness, with 91\% of pages awarded the maximum score; all six dimensions exceed 4.0 on average. Under the current prompts, scientific correctness falls to a mean of 3.54 with a 3\% maximum-score rate, and visual quality dimensions drop below 3.3. A reward concentrated at the top of the scale provides no gradient for further improvement; the revised prompts redistribute scores across the working range, restoring the discrimination that effective RL needs. The Qwen3.5-4B baseline shifts in the same direction at lower absolute values. The interactivity probe and hard-fail gate address a separate class of failure and are treated in Section~\ref{sec:htmlrl}.

\begin{table}[!ht]
  \centering
  \caption{Mean judge scores and ceiling occupancy (share scored 4 or 5 of 5) on 500 pages per model, under the preceding and current judge prompts. Under the current prompts, no visual-quality page reaches 5 on any dimension; content dimensions retain some 4-scores for the stronger model, reflecting genuine quality differences the current anchors can now resolve.}
  \label{tab:reward-dist}
  \small
  \setlength{\tabcolsep}{3.5pt}
  \begin{tabular}{@{\hspace{5pt}}llcccc|cccc@{}}
    \toprule
    \multirow[c]{3}{*}{Judge} & \multirow[c]{3}{*}{Dimension}
      & \multicolumn{4}{c}{Qwen3.5-4B (500 pages)} & \multicolumn{4}{c}{Qwen3.8-27B (500 pages)} \\
    \cmidrule(lr){3-6} \cmidrule(lr){7-10}
    & & \multicolumn{2}{c}{Mean} & \multicolumn{2}{c}{At ceiling} & \multicolumn{2}{c}{Mean} & \multicolumn{2}{c}{At ceiling} \\
    \cmidrule(lr){3-4} \cmidrule(lr){5-6} \cmidrule(lr){7-8} \cmidrule(lr){9-10}
    & & Prec. & Curr. & Prec. & Curr. & Prec. & Curr. & Prec. & Curr. \\
    \midrule
    \multirow[c]{3}{*}{\makecell[l]{Visual\\quality}}
      & layout              & 2.86 & 2.00 & 21\% & 4\%  & 3.84 & 2.92 & 59\% & 29\% \\
      & readability         & 4.12 & 2.94 & 73\% & 28\% & 4.19 & 3.30 & 84\% & 41\% \\
      & aesthetics          & 3.62 & 2.17 & 67\% & 6\%  & 4.49 & 3.28 & 92\% & 47\% \\
    \midrule
    \multirow[c]{3}{*}{Content}
      & instruction fidelity   & 3.37 & 2.54 & 40\% & 4\%  & 4.60 & 3.52 & 94\% & 52\% \\
      & pedagogy               & 2.81 & 2.18 & 30\% & 4\%  & 4.24 & 3.49 & 89\% & 59\% \\
      & scientific correctness & 4.20 & 2.37 & 75\% & 11\% & 4.88 & 3.54 & 97\% & 65\% \\
    \bottomrule
  \end{tabular}
\end{table}

\paragraph{Protocol.}
Better score distributions are a meaningful calibration only if agreement with human judgment is preserved. To verify this, 128 generated pages were independently rated on a 0--5 holistic quality scale---visual design, interaction completeness, pedagogical relevance, and scientific correctness judged jointly, with no access to reward outputs. The sample spans eight prompts with eight rollouts each across two model scales, covering both the weak output of a small pre-RL model and the strong output of a large one. We report Spearman~$\rho$ as the primary metric---rank correlation measures ordinal agreement directly---with Pearson~$r$ as a linear reference.

\begin{table}[!ht]
  \centering
  \caption{Agreement with human ratings on 128 rated pages spanning two model scales, under the preceding and current judge prompts. ``Quality judges only'' includes the visual quality and content judges; ``full pipeline'' additionally applies the viewport defect checks, the interactivity measurement, and the hard-fail gate (18 pages scored zero regardless of visual quality).}
  \label{tab:reward-human}
  \small
  \begin{tabular}{@{\hspace{5pt}}llcc@{\hspace{5pt}}}
    \toprule
    Configuration & Judge prompts & Pearson $r$ & Spearman $\rho$ \\
    \midrule
    \multirow{2}{*}{Quality judges only}   & preceding & 0.609 & 0.672 \\
                                           & current   & \textbf{0.715} & \textbf{0.741} \\
    \midrule
    \multirow{2}{*}{Full pipeline} & preceding & 0.620 & 0.711 \\
                                   & current   & \textbf{0.675} & \textbf{0.749} \\
    \bottomrule
  \end{tabular}
\end{table}

\paragraph{Results.}
Table~\ref{tab:reward-human} quantifies the effect. On the quality-judges-only configuration, the current prompts raise Pearson~$r$ from 0.609 to 0.715 and Spearman~$\rho$ from 0.672 to 0.741. Under the preceding prompts the Qwen3.8-27B baseline averages above 0.84, squeezing nearly all outputs into a narrow band with little ordinal signal; the current prompts restore discrimination, and the correlation improves accordingly. Crucially, this is not a uniform downward shift: anchors that moved all scores together would lower the means without improving correlation, so the gains indicate the prompts now separate pages of different quality.

Adding the full pipeline terms raises agreement further under both prompt versions. For the current prompts, Spearman~$\rho$ increases from 0.741 to 0.749 while Pearson~$r$ moves from 0.715 to 0.675. The rank-agreement gain confirms that the viewport and interactivity signals capture quality dimensions the subjective judges do not fully resolve; the modest Pearson decline reflects a deliberate scoring asymmetry examined below.

\paragraph{Systematic divergence at hard failures.}
Across the rated sample, the reward averages 0.54 against a human mean of 0.25 after normalizing both to $[0,1]$. The overall positive bias reflects the rater's greater use of the lower range. On a specific subset, however, the divergence is directionally reversed and larger in magnitude. Of the 18 pages zeroed by the hard-fail gate, the rater assigned nonzero scores to 13, with a group mean of 0.32 out of 5. These pages render without error and present a complete visual interface, but produce no observable state change in response to user input. A human evaluator applying partial-credit scoring appropriately credits the static components; the gate assigns zero regardless of appearance.

The divergence is by design. The hard-fail gate is not a quality metric but a gradient-suppression mechanism: its purpose is to prevent the policy from collecting reward on non-functional outputs. A policy trained on partial credit for non-responsive pages is incentivized to produce them, a failure mode documented in Section~\ref{sec:htmlrl}. The gate eliminates this incentive at the cost of a predictable disagreement with human scoring on the affected pages. This subset accounts for the majority of the linear-agreement gap between the full-pipeline and quality-judges-only configurations. The gate's trigger conditions and the underlying probe instrumentation are described in Appendix~\ref{app:probe}.

\subsection{External Benchmarks}

We additionally evaluate both modalities on public benchmarks whose contracts
were developed independently of our training suite---testing transfer to
realistic, source-grounded slide briefs and to executable interaction
structures rather than re-scoring with our internal reward.

\paragraph{PresentBench.}
PresentBench contains 238 source-grounded slide-generation instances and an
average of 54.1 instance-specific binary checklist items per instance
\cite{chen2026presentbench}. We render every generated deck to PDF and use
Gemini 3.1 Pro to answer the benchmark's material-independent and
material-dependent checklist questions. Table~\ref{tab:presentbench} reports
the mean and median of the per-instance weighted scores; MI and MD are the
corresponding means for the two checklist groups, and pass rate is the
unweighted fraction of all checklist items answered yes. All four systems
produced a renderable PDF for every instance (238/238). One protocol note: two
checklist calls on the same iPhone source case (one Claude deck and one Luna
deck) repeatedly exceeded the gateway's 120-second limit and were re-run at
the judge's low thinking level; all other calls used the default thinking
level.

\begin{table}[!ht]
  \centering
  \caption{PresentBench results (238 instances, scores in percent). Higher is better.}
  \label{tab:presentbench}
  \small
  \setlength{\tabcolsep}{5pt}
  \begin{tabular}{lrrrrr}
    \toprule
    Model & Mean & Median & MI & MD & Pass rate \\
    \midrule
    \model-27B & 50.48 & 50.52 & 35.35 & 60.56 & 48.55 \\
    \midrule
    Gemini 3.1 Pro & 51.82 & \underline{52.37} & \underline{36.22} & 62.22 & 50.43 \\
    Claude Sonnet 4.6 & \textbf{53.54} & \textbf{54.54} & \textbf{37.16} & \textbf{64.46} & \textbf{52.38} \\
    GPT-5.6 Luna & \underline{51.87} & 52.27 & 36.00 & \underline{62.46} & \underline{50.54} \\
    \bottomrule
  \end{tabular}
\end{table}

The three proprietary models form a narrow leading cluster. Claude Sonnet
4.6 ranks first at 53.54 mean score, while GPT-5.6 Luna (51.87) and Gemini 3.1
Pro (51.82) are nearly tied. \model-27B reaches 50.48, within 1.34 points of
Gemini and 1.39 of Luna despite its much smaller scale; it clears 48.55\% of
all checklist items outright, against 50.4--52.4\% for the proprietary
cluster.

\paragraph{EE-Eval.}
EE-Eval represents each generated explorable explanation as a finite-state
machine (FSM) and compares it with an expert-validated ideal FSM using
structural, semantic, and isomorphism similarities weighted 0.4/0.4/0.2
\cite{wang2026eeeval}.  We run its 127 computer-science topics and supplement
the FSM score with a strict browser check: a page is error-free only if local
headless-Chrome execution raises neither console nor page errors while external
network dependencies are blocked.  Our replication fixes FSM extraction to
GPT-5.6 Terra and uses Qwen3-0.6B embeddings, so its absolute scores should not
be compared directly with those in the original paper.

\begin{table}[!ht]
  \centering
  \caption{EE-Eval replication on 127 topics. Raw is the linear weighted FSM score used for ranking; Display is the benchmark's per-instance nonlinear display transform. Error-free is an independent browser reliability check.}
  \label{tab:ee-eval}
  \small
  \setlength{\tabcolsep}{5pt}
  \begin{tabular}{lrrr}
    \toprule
    Model / configuration & Raw & Display & Error-free \\
    \midrule
    \model-27B-RL-v1 (old reward) & \textbf{57.96} & \underline{93.68} & 116/127 \\
    \model-27B & 57.50 & 92.00 & 123/127 \\
    Claude Sonnet 4.6 (32k repair) & \underline{57.60} & \textbf{94.01} & 123/127 \\
    DeepSeek V4 Flash & 57.59 & 92.30 & \textbf{126/127} \\
    Gemini 3.1 Pro & 57.00 & 89.84 & 120/127 \\
    Claude Sonnet 4.6 (10k) & 52.67 & 89.40 & 79/127 \\
    GPT-5.6 Terra & 49.21 & 85.37 & \textbf{126/127} \\
    GPT-5.6 Luna & 47.31 & 82.13 & \underline{125/127} \\
    \bottomrule
  \end{tabular}
\end{table}

The raw score gives the most faithful cross-model ordering, because the display
transform saturates near 0 and 100 before averaging and can reverse close
comparisons. On that primary measure, \model-27B scores 57.50: 0.46 points
below the old-reward \model-27B-RL-v1, 0.10 below the repaired Claude run, and
0.09 below DeepSeek. The browser check adds a complementary reliability
result: \model-27B is error-free on 123/127 pages, versus 116/127 for
\model-27B-RL-v1 and 126/127 for the
best external systems; all 118 pages on which the generic interaction probe
found a standard control responded successfully. Claude's 32k repair shows why both axes
matter: removing output truncation raises its raw score from 52.67 to 57.60 and
its error-free count from 79 to 123, while its nonlinear display score becomes
the table maximum.

%% file: 6_inference.tex
\section{Inference Acceleration}
\label{sec:inference}

Fast first-pass generation and fast iteration are two different problems. This section describes the two mechanisms that make the \model stack fast at both: \emph{scaffold editing} (Section~\ref{sec:scaffold}) attacks the decoding cost of producing a complete interactive page, while \emph{MAIC-UI} (Section~\ref{sec:maicui}), the authoring harness around the model, attacks the cost of iterating on a page after it exists. The two are orthogonal and compose.

\subsection{Scaffold Editing for First-Pass Generation}
\label{sec:scaffold}

\begin{figure}[!ht]
  \centering
  \includegraphics[width=\linewidth]{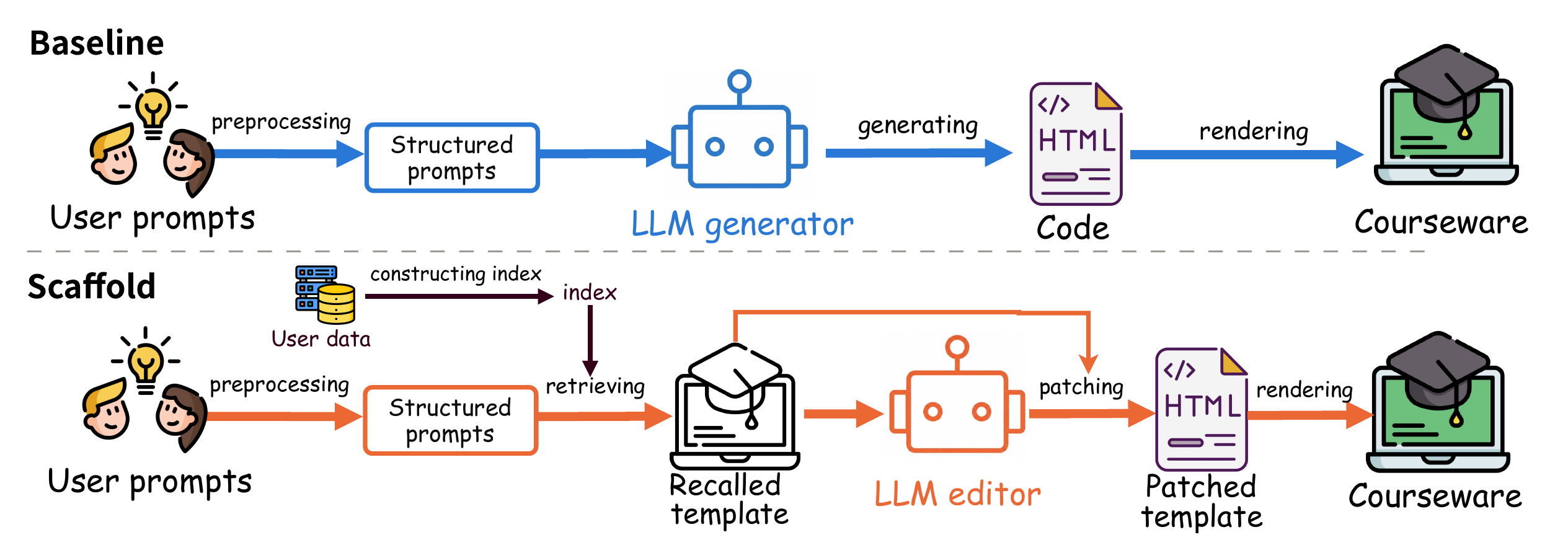}
  \caption{Generation pipelines for interactive courseware. \textbf{Top (baseline)}: user prompts are preprocessed into structured prompts, and the LLM generator emits the complete HTML page from scratch, which is rendered into the courseware. \textbf{Bottom (scaffold)}: an index constructed over accumulated user data grounds retrieval of the closest historical template (\emph{recalled template}); the LLM editor \emph{patches} this template rather than regenerating the page, and the patched template is rendered into the courseware.}
  \label{fig:scaffold-pipeline}
\end{figure}

Producing an interactive page from scratch costs ${\sim}12$k output tokens and ${\sim}67$s, nearly all spent in serial token-by-token decoding. We do not optimize decoding; we eliminate most of it. The key observation: a production system has already accumulated over a million historical widgets, which form a strong structural prior for new requests. \emph{Scaffold editing} therefore reformulates generation as editing---retrieve the most similar historical template, let the model emit only component-level edit decisions, and reconstruct the page programmatically (Figure~\ref{fig:scaffold-pipeline}).

\paragraph{Approach.}
The pipeline has three stages. \textbf{(i) Retrieval}: over the 1M-template corpus, a semantic search returns the closest historical template, with a tiered router that falls back to direct generation whenever the retrieved scaffold is a poor match---so the method never degrades below baseline. \textbf{(ii) Component-level decisions}: the template is decomposed into a component tree (configuration, styles, body, and each JS function separately); the model emits one \textsc{keep}/\textsc{modify}/\textsc{replace} decision per component under a strictly lazy policy---keep what can be kept, patch what can be patched, rewrite only what must change. Splitting JS at function granularity is the single largest lever: the model writes decisions only for functions that actually change, which alone removed $74\%$ of output tokens in a representative sample. \textbf{(iii) Programmatic reconstruction}: decisions are applied deterministically by code, never re-typed by the model---unmodified components cost zero tokens and carry zero transcription risk, and a per-component demotion chain (failed MODIFY $\rightarrow$ REPLACE $\rightarrow$ KEEP) confines any failure to a single component.

\begin{table}[!ht]
  \centering
  \caption{Scaffold editing vs.\ direct generation. Bold marks the deployed scaffold results; the oracle row (retrieving each sample's own template) is an upper bound, not a deployable configuration. External rows use a 500-request test set generated independently of the corpus; both scaffold rows run without a routing threshold and include auto-fallback costs end-to-end.}
  \label{tab:scaffold}
  \small
  \begin{tabular}{llcccc}
    \toprule
    Test set & System & Output (tok) & Latency avg/p90 (s) & $\Delta$ tokens & Success \\
    \midrule
    \multirow{2}{*}{Internal} & Direct & 12,344 & 67.2 / 89.6 & --- & --- \\
     & Scaffold & \textbf{2,999} & \textbf{16.0 / 28.1} & \textbf{$-$75.7\%} & 94\% \\
    \cmidrule(lr){1-6}
    Internal, oracle & Scaffold & 2,161 & 12.7 / 22.5 & $-$82.5\% & --- \\
    \cmidrule(lr){1-6}
    \multirow{2}{*}{External} & Direct & 8,468 & 50.9 / 75.9 & --- & --- \\
     & Scaffold & \textbf{3,411} & \textbf{24.7 / 43.5} & \textbf{$-$59.7\%} & 98\% \\
    \bottomrule
  \end{tabular}
\end{table}

\begin{figure}[!ht]
  \centering
  \begin{subfigure}[t]{0.67\linewidth}
    \includegraphics[width=\linewidth]{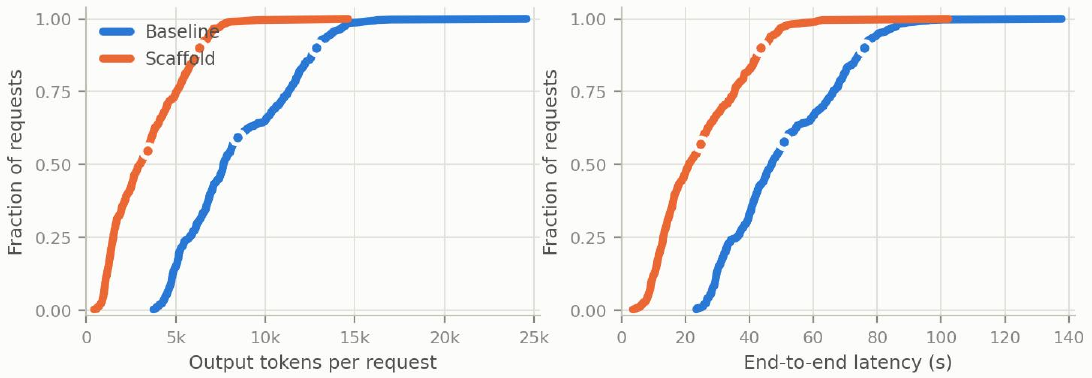}
    \caption{Per-request output tokens (left) and end-to-end latency (right); dot markers denote the mean and p90.}
    \label{fig:ecdf-external}
  \end{subfigure}\hfill
  \begin{subfigure}[t]{0.31\linewidth}
    \includegraphics[width=\linewidth]{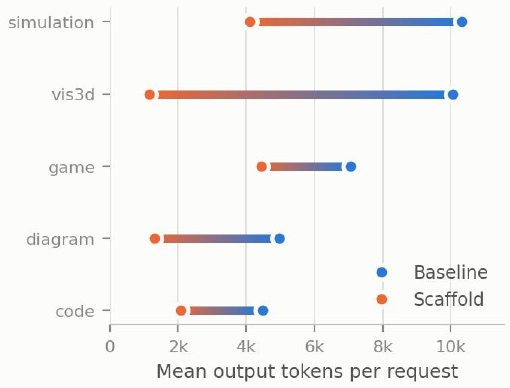}
    \caption{Mean output tokens by widget type.}
    \label{fig:subtype-external}
  \end{subfigure}
  \caption{External test set ($n{=}500$ per arm); the scaffold arm is charged end-to-end, including 11 auto-fallback runs. The scaffold distribution lies left of baseline at every percentile in (a); per-type reductions in (b) span $-37\%$ (game) to $-88\%$ (vis3d), with code and vis3d using an expanded subsample ($n{=}50$ per arm per type)---structure-heavy templates are reused almost wholesale, whereas game logic forces more component rewrites.}
  \label{fig:external-dists}
\end{figure}

\paragraph{Results.}
Table~\ref{tab:scaffold}: on the internal test set, output falls to 2,999 tokens and latency to 16.0s (${\sim}-76\%$ both axes) at a 94\% reconstruction success rate, against an oracle bound of $-82.5\%$ (retrieving each sample's own template). On an external test set sharing nothing with the corpus, gains remain $-60\%$ tokens / $-52\%$ latency, and the reduction holds at every percentile of the per-request distribution (Figure~\ref{fig:ecdf-external})---evidence that the acceleration generalizes beyond near-duplicates. The honest boundary: gains are a function of corpus coverage (Figure~\ref{fig:subtype-external}: per-type reductions span $-37\%$ for game to $-88\%$ for vis3d); topically related but structurally distant templates (similarity 0.85--0.93) can force near-baseline output. Quality is guarded by hard gates (render/runtime failure $\Rightarrow$ fail) plus automatic interaction probes, so the reported gains are grounded in objective, end-to-end measures.

\subsection{Fast Iterative Editing: Click-to-Locate in the MAIC-UI Harness}
\label{sec:maicui}

\begin{figure}[!ht]
  \centering
  \includegraphics[width=\linewidth]{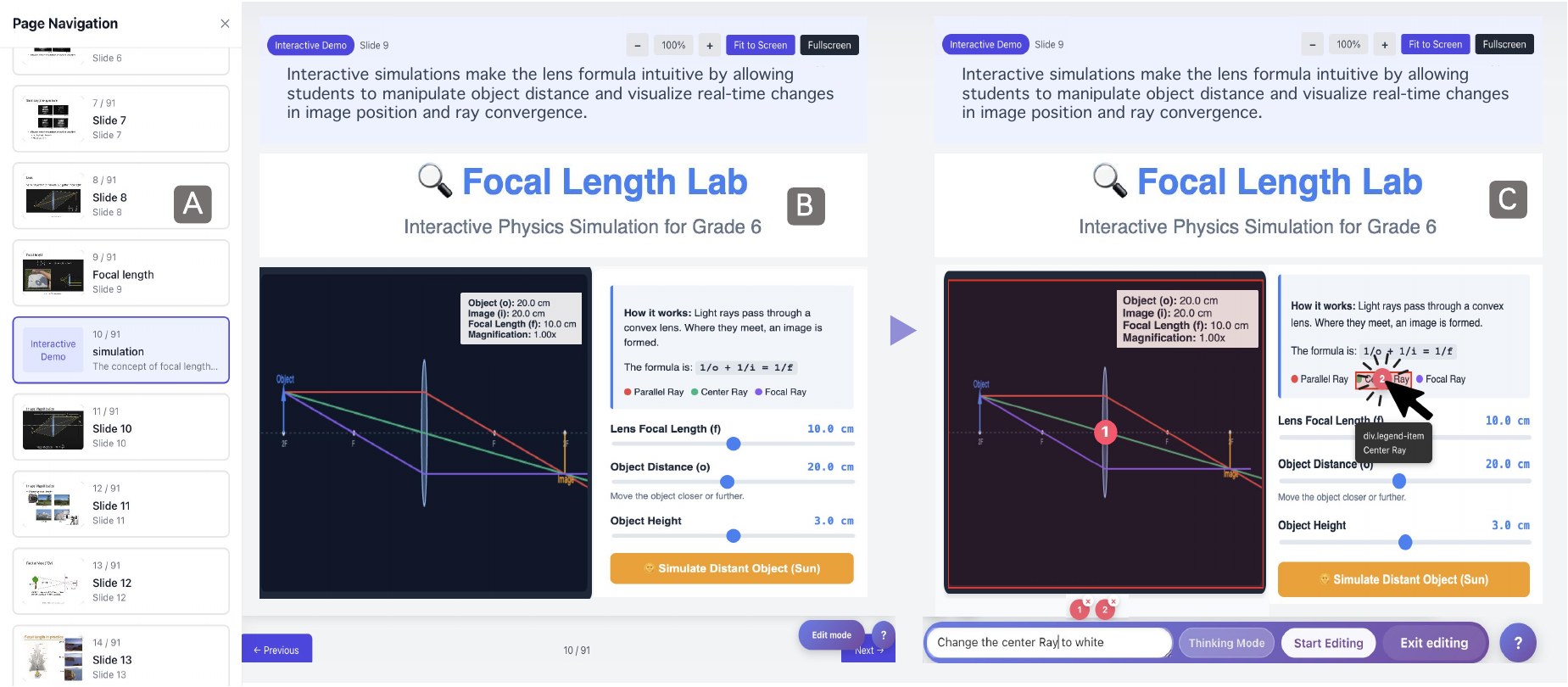}
  \caption{The MAIC-UI authoring harness. The Click-to-Locate editing loop (right): clicking an element in the live preview anchors the edit via its DOM context; the model returns a unified diff applied incrementally.}
  \label{fig:maicui-system}
\end{figure}

The cost of interactive courseware is dominated not by the first generation but by iteration: teachers request small changes---reword a title, fix a formula, adjust a parameter range---and each request historically triggers full-file regeneration. Reported systems take 200--600 seconds per such edit~\cite{fakhoury2024llm,huang2024new}. MAIC-UI~\cite{tu2026maicuimakinginteractivecourseware}, the authoring harness shipped with OpenMAIC (our primary application partner, with whom all testing and production application were carried out jointly) and released with \model-4B, reduces the average edit to seconds. The enabling technique is \emph{Click-to-Locate incremental editing} (Figure~\ref{fig:maicui-system}), which combines three ingredients.

\paragraph{Click-to-Locate element anchoring.}
Asking teachers to describe which element to change is ambiguous; asking them to navigate source code is unrealistic. Instead, the frontend embeds a Web-Inspector-style citation system: clicking any element in the live preview captures its XPath and CSS selector~\cite{benedikt2009xpath} and displays the anchored HTML snippet with a highlight overlay. Point-and-click replaces both natural-language localization and code navigation. Beyond usability, the captured DOM context is also what makes precise patch anchoring reliable: it supplies the exact local structure that the edit target must match against.

\paragraph{Unified-diff incremental generation.}
Given the anchored element and a natural-language instruction, the model returns a \emph{unified diff}~\cite{nugroho2020different} rather than a regenerated file: only changed lines plus minimal context, a ${\sim}90\%$ reduction in output tokens compared with full-file regeneration. Diffs are applied client-side with fuzzy context matching that absorbs minor formatting drift between the model's expected context and the live DOM. In deployed classroom scenarios, typical element edits complete in under 10 seconds.

\paragraph{Lean, task-aligned context.}
The editing prompt carries only the selected element, the instruction, and the necessary page context---thousands of tokens, not the general-purpose programming environment that coding agents load on every interaction. This is the difference between an application-layer harness and a general agent: the context is specialized for ``edit this page'' rather than ``program anything.''

\begin{figure}[!ht]
  \centering
  \begin{subfigure}[b]{0.49\linewidth}
    \includegraphics[width=\linewidth]{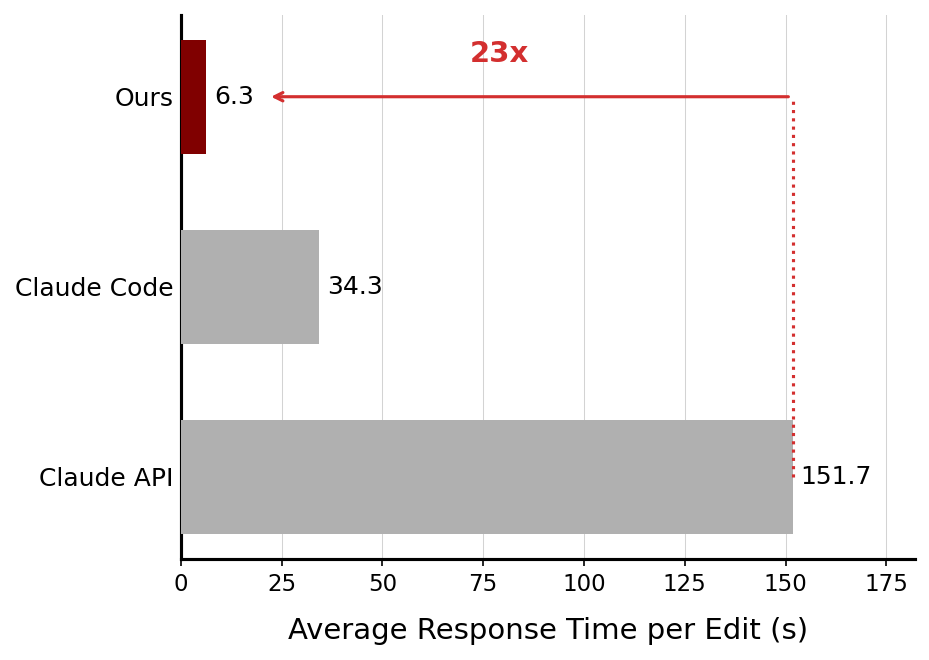}
    \caption{Response time per edit}
  \end{subfigure}
  \hfill
  \begin{subfigure}[b]{0.49\linewidth}
    \includegraphics[width=\linewidth]{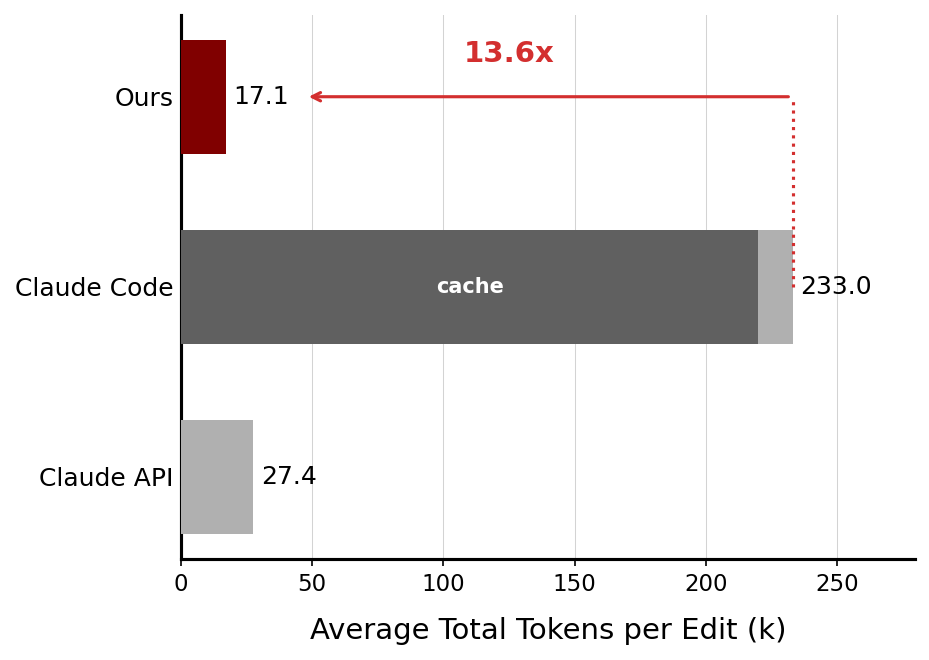}
    \caption{Tokens per edit}
  \end{subfigure}
  \caption{Editing efficiency on the same backbone model across three editing tasks: MAIC-UI vs.\ Claude Code vs.\ direct API regeneration.}
  \label{fig:maicui-eff}
\end{figure}

\paragraph{Editing-efficiency study.}
To isolate the effect of harness architecture from model capability, we compared three editing stacks over the same three editing tasks on the same backbone model: (i) MAIC-UI; (ii) Claude Code, a general-purpose coding agent; and (iii) a direct-API loop that regenerates the file per request. MAIC-UI completes an average edit in 6.3\,s with 17.1k tokens at ¥0.37, versus 34.3\,s / 233k tokens (94\% cache-hit) / ¥1.03 for Claude Code and 151.7\,s / 27.4k tokens / ¥1.68 for the direct API (Figure~\ref{fig:maicui-eff}). The $23\times$ latency reduction against direct regeneration---and $5\times$ against a general-purpose agent, even crediting its full caching advantage---comes from application-layer specialization: a lean, task-aligned context (17.1k tokens per edit) instead of a general-purpose programming environment, and DOM-anchored diffs instead of whole-file rewrites.

%% file: 7_deployment.tex
\section{Deploying \model on Domestic Accelerators}
\label{sec:deployment}

Adapting \model to domestic accelerators is what makes the cost story of Section~\ref{sec:intro} real: it removes the dependency on premium GPUs for serving. The challenge is that \model-27B is exactly the kind of architecture that new hardware struggles to support---a hybrid interleaving 48 gated-delta-net (GDN) linear-attention layers~\cite{yang2024gated} with 16 full-attention layers~\cite{dao2022flashattention}, shipped as an FP8 checkpoint for an arithmetic the target hardware does not implement, on an engine release that registers no NPU backend for the architecture at all. We use a single Ascend 910 A3 node as a case study. Four findings carry the section: three make the deployment work; the fourth explains what cannot work yet, and why.

\paragraph{Cross-precision reconstruction, verified layer by layer.}
We dequantize the FP8 weights to BF16 and re-quantize all 256 Linear layers to INT8 W8A8 under per-channel symmetric scaling~\cite{jacob2018qat,ocp2023mx}, holding tensors within ${\sim}1\%$ of their originals. Reconstructions of this kind can silently alter the computed function while outputs remain fluent, so we verify per tensor with \textbf{Noise-Floor Arbitration}: each tensor's admissible deviation is derived from its own measured BF16 rounding floor against an FP64 reference---a global tolerance would admit or reject everything at once. Under this criterion the full 64-layer prefill path matches the reference, including all 16 paged-attention layers; fused-add and INT8 GEMM paths are bitwise identical.

\paragraph{Release-independent dispatch, cleanly attributed.}
Vendor support for this architecture arrived incrementally across engine releases, so waiting for upstream support is not a schedulable strategy. Our \textbf{Attention Dispatch Override} turns kernel selection into a deployment-time decision: linear-attention layers are rerouted from a defective default operator to the fused-infer-attention operator, with a chunked cache replacing the incompatible Mamba radix cache. On a release predating upstream support, this reroute alone serves at production validity. The reroute is structural, not a stopgap: graph capture runs only on the operator it selects, and a cross-release control (Table~\ref{tab:ascend-attribution}) attributes the entire \textbf{1.99$\times$} throughput gain to graph-capture-enabled execution, not to the engine upgrade.

\begin{table}[!ht]
  \centering
  \caption{Cross-release attribution control (single die, concurrency 6). The later release with graph capture disabled reproduces the reroute baseline to 0.2\%, so the full $1.99\times$ is attributable to graph capture. Greedy quality: 30/30 valid business cases in all three arms.}
  \label{tab:ascend-attribution}
  \small
  \begin{tabular}{lcc}
    \toprule
    Configuration & RPM/die & Relative \\
    \midrule
    v0.5.9 (reroute baseline) & 1.885 & 1.00$\times$ \\
    v0.5.16, graph capture disabled & 1.881 & 1.00$\times$ \\
    v0.5.16, graph capture enabled & \textbf{3.743} & \textbf{1.99$\times$} \\
    \bottomrule
  \end{tabular}
\end{table}

\paragraph{Replica-first decomposition.}
Under a fixed die budget, the conventional heuristic maximizes tensor parallelism~\cite{megatron} to maximize KV capacity. For this hybrid model the heuristic inverts: raising TP enlarges running-request capacity from 24 to 192 slots while throughput falls monotonically---the binding resource is intra-TP communication, not memory. We reduce TP to the smallest degree that admits the BF16 weights on 64\,GB dies (TP2) and spend the remaining dies on independent replicas: \textbf{31\%} higher throughput than the capacity-maximizing topology at matched cache hit. A latency knee at concurrency 64 (beyond it, throughput +34\% while TTFT degrades 36$\times$) sets the interactive operating point.

\paragraph{The recurrent-state constraint---and how to catch its silent failures.}
The deeper finding is a single root cause behind two unavailable accelerations. Both prefix caching and speculative decoding require reconstructing the recurrent state at a position other than where it was produced. For attention layers this is trivial---the KV cache is an append-only, position-indexed log~\cite{vllm}, so truncation is an information-preserving pointer operation. The GDN state, however, evolves as
$S_t = g_t \odot S_{t-1} + \beta_t k_t v_t^{\top}$ with decay gate $g_t \in (0,1)$: the update is a contraction, admits no inverse, and \emph{without an inverse there is no truncation}. On this stack both mechanisms fail silently---outputs stay fluent and well-formed while being wrong: a warm prefix-cache run re-emits the tail of the prompt (\emph{Hot--Cold Divergence}, HCD), and speculative verification diverges from sequential decoding from the second position onward despite a healthy draft head (\emph{Bitwise Speculative Equivalence}, BSE). Both tests are cheap, require no reference implementation, and apply to any stack that reuses recurrent state; we contribute them as operator-acceptance tests for state side effects, which forward-output comparison cannot catch. Where state must be materialized, replay beats snapshotting by \textbf{160$\times$} in per-request scratch (3.8\,MB vs.\ 604\,MB), a difference that competes directly with the KV budget on a 64\,GB die.

\paragraph{Outcome: application-level parity, gap honestly decomposed.}
Table~\ref{tab:ascend-e2e} compares the adapted stack against the production A800 deployment of the same model family on the same 500-case business dataset: \textbf{500/500} valid parses on both, mean element counts 17.4 vs.\ 17.4, mean output lengths 1742.7 vs.\ 1749.7 tokens---indistinguishable at the application level. The remaining end-to-end gap decomposes multiplicatively into $3.63\times = 1.48\times$ (hardware and software stack) $\times\, 2.45\times$ (speculative decoding unavailable)---and the larger factor is not a property of the accelerator: it traces to the operator defect above plus a single-speculative-layer checkpoint, not to numerical precision. Placing precision, kernel binding, and decomposition under deployment-time control decouples deployment readiness from the vendor's release cadence---which is what lets a new architecture become schedulable on a rapidly evolving accelerator ecosystem.

\begin{table}[!ht]
  \centering
  \caption{End-to-end comparison on the 500-case business dataset (same client, natural stopping). The gap decomposition follows consecutive rows: A800 production $\rightarrow$ A800 without speculative decoding isolates the $2.45\times$; A800 without speculative decoding $\rightarrow$ Ascend at the c64 operating point isolates the $1.48\times$.}
  \label{tab:ascend-e2e}
  \small
  \begin{tabular}{llllcc}
    \toprule
    Platform & Topology & Weights & Spec.\ decoding & RPM & Parse \\
    \midrule
    A800 (production) & TP2$\times$DP4 & FP8 W8A8 & 4-step & 167.4 & 500/500 \\
    A800 & TP1$\times$DP8 & FP8 W8A8 & --- & 68.4 & 500/500 \\
    \textbf{Ascend 910 A3} & TP2$\times$DP8 & INT8 W8A8 & --- & \textbf{46.1}\,(c64) & 500/500 \\
    Ascend 910 A3, peak & TP2$\times$DP8 & INT8 W8A8 & --- & 70.4\,(c128) & 500/500 \\
    \bottomrule
  \end{tabular}
\end{table}


%% file: 9_conclusion.tex
\section{Conclusion}
\label{sec:conclusion}

We presented \model, a family of post-trained models for Learning Environment Generation, together with the evaluation suites, rewards, and serving infrastructure that make the task measurable and cheap. Two findings generalize beyond this system. First, \emph{interactivity must be measured, not judged}: every reliability gain in this report traces to executable probes in the reward loop, and the one failure we disclose---a checkpoint that scored highest on code while its games were unplayable---is exactly what a screenshot-only judge cannot see. Second, \emph{what the reward cannot measure, RL will quietly destroy}; reward design, not data volume, set the ceiling of every stage here. \model-27B serves production traffic---deployed jointly with the OpenMAIC team, our primary application partner---at a fraction of flagship cost, and the full stack runs on domestic accelerators. We release \model-4B openly, weights and editing harness together, so that efficient, reliable generation of learning environments can reach the classrooms that need it most.

%% file: appendix.tex
\appendix

\input{8_future}

\section{Judge Prompts}
\label{app:prompts}

Three prompts define every score reported in this paper. The slide judge (Gemini 3.1 Pro) scores fidelity and layout for slide-std / slide-short. The HTML visual-quality and content prompts are judged by a 27B VLM inside the hardened reward; together with the interactivity probe and the dual-viewport checks they form the HTML composite. The prompts are internal to keep the scoring pipeline---which our RL reward shares---fixed and confidential; external models are scored through the same pipeline (Section~\ref{sec:eval}).



\section{Interactivity Measurement}
\label{app:probe}

Section~\ref{sec:rl} claims that interactivity must be measured rather than judged. This appendix describes the instrument that does the measuring and the gate built on top of it.

\paragraph{Observability Limits of Static Rendering.}

Every other term in the HTML reward reads a static rendering. That is adequate for layout and readability, which are properties of a frame, and inadequate for interactivity, which is a property of a page's response to input and therefore invisible in any single frame. The limitation has direct consequences: a page whose opening frame is well composed and whose controls are inert scores well on every screenshot-based dimension, and a policy optimizing those dimensions alone will find that region of output space. Section~\ref{sec:htmlrl} records it doing so.

Closing this gap requires operating the page directly. Each candidate is rendered in a Playwright-driven Chromium instance that enumerates the page's interactive elements (buttons, range inputs, selects, canvases, drag targets) and drives them, then reports what changed. Two properties of the harness matter for the reward: it must not credit a page for changes it would have produced anyway, and it must not penalize a page for controls the harness itself is too blunt to operate.

\paragraph{Detecting response.}

The first generation of the probe compared a DOM fingerprint (a hash over visible text and layout geometry) before and after each synthetic event, and sampled WebGL canvases through \texttt{gl.readPixels} to catch view changes that leave no DOM trace. This is sound for pages whose state lives in the DOM and systematically wrong for the sub-type that matters most. A physics simulation keeps its state inside a canvas; its surrounding DOM often updates on a timer regardless of input. The fingerprint sees the timer, reports a response, and a page whose canvas ignores every control passes.

The current generation instruments the drawing itself. Wrapping the Canvas 2D entry points, including \texttt{fillRect}, \texttt{drawImage}, \texttt{stroke}, and \texttt{fill}, lets the probe record a \emph{signature} per drawing call, composed of operation type, coordinates, and color, and maintain the set of signatures observed. Response is then the arrival of signatures that were not in the set before the interaction---a statement about the canvas itself, not about its surroundings.

Self-running animation would defeat this on its own, since an animating canvas emits new signatures continuously whether or not anyone touches it. The probe therefore holds the page quiescent for a fixed interval before interacting and records the signatures that appear unprompted; that baseline is subtracted, and only the excess counts as response. WebGL canvases, which produce no 2D draw calls, are handled by comparing sampled framebuffer regions across an input differential and separately checking that the scene is not frozen.

Four signals emerge: registered listener count, post-baseline signature delta, per-control response rate over the traversal, and WebGL activity. The reward consumes their coverage-weighted mean rather than any one of them, so that a page is credited in proportion to how much of its interactive surface the probe could confirm working.

\paragraph{Traversal strategy.}

Naive traversal under-reports on two sub-types, both because the page's interesting state is gated behind a control the probe must find first. Simulations commonly open in a standby state with an empty canvas until a start control is pressed; capturing or probing before that yields a blank frame that the visual judges would correctly, and uselessly, score as broken. Diagrams built as stepped walkthroughs show only their first stage until advanced. The probe therefore identifies start-like and step-like controls before general traversal and presses them, stepping a walkthrough to its final stage, and only then captures the frame the judges will see and begins measuring the remaining controls. This is why the screenshots reaching the visual judges depict an initialized page, and why a blank canvas surviving to that point is strong evidence rather than a timing artifact.

Some controls resist automation for reasons that say nothing about page quality: dragging a card into a target slot, or completing an ordered multi-step gesture, requires semantic understanding the probe does not have. These are recognized, left unoperated, and excluded from the response statistics so that a game built around such a mechanic is not recorded as unresponsive merely because the harness could not play it.

The whitelist is not a permanent ceiling. Two directions are worth pursuing. The first is richer action primitives: targeted drag-and-drop, gesture sequences, and typed input that satisfies a field's expected format would shrink the whitelist without requiring the harness to understand the page. The second direction is more consequential. Replacing scripted traversal with a lightweight agent that reads the page as a learner would, choosing controls based on their labels and pedagogical context and verifying that what changed constitutes a meaningful response, would close the gap between mechanical coverage and genuine interactivity assessment. Both directions face the constraint that makes the current probe practical. As measured under the HTML RL workload in Appendix~\ref{app:reward-server-perf}, reward computation on a 4B-generated page averages 55 seconds per GRPO step at batch size 64, and the P99 stays within 76 seconds. A general-purpose web agent capable of semantic page operation takes five minutes to tens of minutes per page---roughly an order of magnitude slower and incompatible with per-step reward latency requirements. That gap sets the engineering challenge, and closing it is where future work on interactive-page evaluation has the most to offer.

\paragraph{The hard-fail gate.}

The graded signal above enters the reward as one weighted term among five, which is the right treatment for partial unresponsiveness. It is the wrong treatment for total unresponsiveness. A page that cannot be entered has no educational value at any level of visual polish, yet a fully inert page whose judged dimensions all score respectably still lands mid-range once its interactivity defect is averaged in at weight $0.3/1.2$. Mid-range is a direction a policy will happily climb. Four conditions therefore bypass the weighted mean and assign zero outright.

The first is a page whose every probe-operable control was driven and produced no observable change in DOM, canvas signatures, or WebGL state. Whitelisted controls are excluded, so the condition fires only on confirmed inertness rather than on harness limitations. The second and third target blank canvases: a simulation whose canvas remains a large uniform field after its start control has been pressed, and the same condition observed on the tablet or mobile rendering, which catches pages that initialize on desktop and collapse at narrower widths. The fourth is narrower and addresses the failure that motivated the gate: a game whose single operable control never changes the page state, the signature of a title screen with a dead start button.

The gate is a mechanism for making a gradient unavailable, not a quality measurement, and it deliberately scores some pages far below what a human grader would give them. Section~\ref{sec:rewardhuman} quantifies that divergence and argues for keeping it.

\section{HTML RL Reward Server}
\label{app:reward-server-perf}

The reward server described in Section~\ref{sec:rewarddesign} handles all scoring during HTML RL training. Each \texttt{/reward} request passes through two phases before returning. In the first phase, one of four Playwright-driven Chromium render workers executes the page and captures the screenshots and interactivity probe trace. In the second phase, the rendered outputs are dispatched concurrently to the VLM judge and the deterministic viewport pipelines. Timing data from the benchmark runs reveals where latency is spent: rendering averages 7.5 seconds for Qwen3.5-4B HTML and 13 seconds for Qwen3.8-27B HTML, with P95 values of 12 and 21 seconds respectively. The VLM judge calls that follow---rubric and fidelity running in parallel---average 33--38 seconds for 4B HTML and 60--63 seconds for 27B HTML, accounting for the majority of end-to-end latency in both cases. Qwen3.8-27B-generated pages induce roughly double the judge time of 4B pages, reflecting their greater length and structural complexity. The measurements in this appendix characterize the full round-trip latency under both the RL and evaluation workloads, and compare the two candidate judge models on throughput and scoring consistency.

\paragraph{Workload and concurrency.}
The slime GRPO framework issues reward requests in synchronous batches: every RL step generates eight rollouts for each of eight prompts, placing exactly 64 concurrent requests at the server. The training loop waits for the full batch to return before computing advantages for the next step. Under double-buffering, where rollout generation for the next step overlaps with reward scoring for the current one, up to 128 requests may be in flight simultaneously. The admission semaphore is set to 160 to accommodate this without queuing. Each request spawns up to four parallel judge calls, so the effective peak concurrency into the VLM judge reaches approximately $160 \times 4 = 640$; the judge serving configuration is calibrated for this load. The evaluation workload differs in pattern: a semaphore of 64 keeps that many requests continuously in flight without batch boundaries, approximating the evaluation harness and the online serving pattern.

\paragraph{Throughput and latency.}
The HTML-500 benchmark pages from Section~\ref{sec:eval}---shuffled at a fixed seed so that pages of the same sub-type are not batched together, which would otherwise skew latency measurements by creating bursts of uniformly fast or slow requests---were scored three times under each workload to measure per-request latency and total round time. Results are shown in the table below, which covers both judge models and shows that the two are within 5\% of each other on every latency metric. Per-step reward latency under the RL workload is the round time divided by the number of steps: $441\text{--}444\,\text{s} / 8 \approx 55\,\text{s}$ for Qwen3.5-4B HTML, and $644\text{--}648\,\text{s} / 8 \approx 81\,\text{s}$ for Qwen3.8-27B HTML. Both lie within the 120-second step budget measured during training, leaving margin for gradient computation. The P99 tail reaches 76--82 seconds but does not block the step, because the training loop waits for the full batch rather than individual requests. To contain the long tail, each judge call uses a hedge strategy: after 32 seconds without a response, a duplicate request is dispatched to a second worker; whichever reply arrives first is accepted and the other discarded. This keeps stalled requests from accumulating to the full timeout and ensures P99 stays within 82 seconds even under an uneven vLLM queue.

\begin{table}[H]
  \centering
  \caption{Reward server latency, fallback rate, and round time on 500 shuffled pages, three runs per configuration. Latency is per-request wall-clock time. Under the RL workload with batch size 64, a round of 500 pages spans $500/64 \approx 8$ GRPO steps.}
  \label{tab:reward-perf}
  \small
  \setlength{\tabcolsep}{3.5pt}
  \begin{tabular}{@{\hspace{5pt}}lllrrrrrrr@{\hspace{5pt}}}
    \toprule
    Judge & HTML & Workload & Mean & Median & P95 & P99 & Fallback & Round \\
    \midrule
    \multirow{4}{*}{Qwen3.8-27B}
      & \multirow{2}{*}{Qwen3.5-4B}  & RL   & 38\,s & 38\,s & 71\,s & 76\,s & 0.27\% & 441\,s \\
      &                               & Eval & 57\,s & 71\,s & 92\,s & 96\,s & 0.00\% & 456\,s \\
      & \multirow{2}{*}{Qwen3.8-27B} & RL   & 68\,s & 76\,s & 80\,s & 82\,s & 0.33\% & 648\,s \\
      &                               & Eval & 70\,s & 78\,s & 87\,s & 89\,s & 0.00\% & 563\,s \\
    \midrule
    \multirow{4}{*}{Qwen3.6-27B}
      & \multirow{2}{*}{Qwen3.5-4B}  & RL   & 40\,s & 42\,s & 71\,s & 75\,s & 0.33\% & 444\,s \\
      &                               & Eval & 57\,s & 70\,s & 91\,s & 95\,s & 0.00\% & 454\,s \\
      & \multirow{2}{*}{Qwen3.8-27B} & RL   & 67\,s & 75\,s & 79\,s & 81\,s & 0.47\% & 644\,s \\
      &                               & Eval & 69\,s & 78\,s & 87\,s & 90\,s & 0.00\% & 559\,s \\
    \bottomrule
  \end{tabular}
\end{table}

A fallback occurs when all judge calls for a given pipeline time out or return an error, in which case that pipeline contributes a neutral default score rather than a measured one and the request is flagged for downstream filtering. Fallback rates are near zero across both judges and both workloads; the eval workload reaches exactly zero in every run. Hard-fail rates reflect model capability rather than server load: Qwen3.5-4B HTML triggers the gate on approximately 22.5\% of pages, while Qwen3.8-27B HTML triggers it on 12.5\%.

\paragraph{Scoring robustness.}
The same 500 pages were scored three times per judge under the RL workload, and the per-page reward span over three runs was computed for each judged dimension. The table below compares Qwen3.8-27B and Qwen3.6-27B side by side. Viewport and interactivity dimensions show near-identical consistency between the two judges. The viewport pipelines use a 0--2 severity scale with a fixed defect checklist graded by the VLM---each criterion either triggers or does not, leaving little room for between-run variation---which is why their exact-agreement rates reach 92--98\%. Interactivity is a deterministic probe measurement and varies only with the page itself. The subjective dimensions show a consistent advantage for Qwen3.8-27B, particularly on scientific correctness, where its exact-agreement rate is 56\% versus 46\% for Qwen3.6-27B and its mean span is 0.55 versus 0.67.

\begin{table}[H]
  \centering
  \caption{Per-dimension scoring consistency across three independent RL-workload runs on 500 Qwen3.5-4B pages, comparing Qwen3.8-27B and Qwen3.6-27B as judges. Exact is the fraction of pages receiving identical scores in all three runs. Visual quality and content dimensions use a 0--5 integer scale; viewport dimensions use 0--2; interactivity defect is continuous in $[0,1]$.}
  \label{tab:reward-robust}
  \small
  \setlength{\tabcolsep}{5pt}
  \begin{tabular}{@{\hspace{5pt}}llc cc cc@{\hspace{5pt}}}
    \toprule
    \multirow[c]{2}{*}{Pipeline} & \multirow[c]{2}{*}{Dimension} & \multirow[c]{2}{*}{Scale}
      & \multicolumn{2}{c}{Qwen3.8-27B} & \multicolumn{2}{c}{Qwen3.6-27B} \\
    \cmidrule(lr){4-5} \cmidrule(lr){6-7}
    & & & Exact & Span & Exact & Span \\
    \midrule
                                & layout               & 0--5    & 81\% & 0.20 & 75\% & 0.29 \\
    \makecell[l]{Visual quality} & readability        & 0--5    & 67\% & 0.34 & 61\% & 0.42 \\
                                & aesthetics           & 0--5    & 67\% & 0.35 & 61\% & 0.42 \\
    \midrule
                                & instruction fidelity & 0--5    & 77\% & 0.23 & 71\% & 0.33 \\
    Content                     & pedagogy             & 0--5    & 65\% & 0.38 & 61\% & 0.42 \\
                                & scientific correctness & 0--5  & 56\% & 0.55 & 46\% & 0.67 \\
    \midrule
                                & missing element      & 0--2    & 92\% & 0.12 & 92\% & 0.12 \\
    \makecell[l]{Viewport (tablet)} & new occlusion   & 0--2    & 95\% & 0.09 & 96\% & 0.07 \\
                                & scaling              & 0--2    & 98\% & 0.04 & 97\% & 0.06 \\
    \midrule
                                & missing element      & 0--2    & 91\% & 0.12 & 91\% & 0.13 \\
    \makecell[l]{Viewport (mobile)} & new occlusion   & 0--2    & 93\% & 0.09 & 89\% & 0.14 \\
                                & scaling              & 0--2    & 91\% & 0.15 & 86\% & 0.24 \\
    \midrule
    Interactivity               & defect               & $[0,1]$ & 93\% & 0.02 & 93\% & 0.01 \\
    \bottomrule
  \end{tabular}
\end{table}

\paragraph{Judge model selection.}
Two candidate judge models---Qwen3.8-27B and Qwen3.6-27B---were evaluated on the same hardware under identical serving configurations. Their throughput and latency are virtually indistinguishable, differing by less than 5\% on mean latency and at most 2 percentage points on any fallback rate across all workload and HTML-model combinations. The choice therefore rests on alignment with human judgment and on dimension-level scoring consistency.

Agreement with the human ratings collected in Section~\ref{sec:rewardhuman} was re-measured under both judge models on the same 128 annotated pages, spanning two model scales and eight prompts each, under two judge-prompt versions. Numbers are reported in the table below. Under the current judge prompts with quality judges only, Qwen3.8-27B achieves Spearman $\rho = 0.741$ against Qwen3.6-27B's 0.716. The advantage is concentrated in text-intensive dimensions: Qwen3.8-27B's mean span for scientific correctness is 0.46 versus 0.55 for Qwen3.6-27B under the RL workload, reflecting more consistent comprehension of factual claims. The Qwen3.6-27B full-pipeline $\rho$ of 0.773 slightly exceeds Qwen3.8-27B's 0.749, but this reversal traces to lower viewport variance on complex pages rather than to better subjective scoring. Qwen3.8-27B is selected as the production judge on the basis of its higher alignment under the quality-judges-only configuration and its substantially lower variance on scientific correctness, the dimension most dependent on deep language understanding.

\begin{table}[H]
  \centering
  \caption{Agreement with human ratings on 128 annotated pages under two judge-prompt versions. Quality judges only includes the visual quality and content pipelines; full pipeline additionally applies viewport defect checks, the interactivity measurement, and the hard-fail gate.}
  \label{tab:judge-iaa}
  \small
  \begin{tabular}{@{\hspace{5pt}}llllcc@{\hspace{5pt}}}
    \toprule
    Judge & Prompts & Configuration & $n$ & Pearson $r$ & Spearman $\rho$ \\
    \midrule
    \multirow{4}{*}{Qwen3.8-27B}
      & preceding & quality judges only & 128 & 0.609 & 0.672 \\
      & current   & quality judges only & 128 & \textbf{0.715} & \textbf{0.741} \\
      & preceding & full pipeline       & 128 & 0.620 & 0.711 \\
      & current   & full pipeline       & 128 & 0.675 & 0.749 \\
    \midrule
    \multirow{4}{*}{Qwen3.6-27B}
      & preceding & quality judges only & 128 & 0.644 & 0.687 \\
      & current   & quality judges only & 128 & 0.687 & 0.716 \\
      & preceding & full pipeline       & 128 & 0.619 & 0.713 \\
      & current   & full pipeline       & 128 & 0.680 & \textbf{0.773} \\
    \bottomrule
  \end{tabular}
\end{table}

\section{System Prompts}
\label{app:sysprompts}
The system prompt is the task interface: it selects the modality and carries the output contract (Section~\ref{sec:task}). Training used exactly six---the slide contract and five interactive-HTML templates, one per sub-type (learning pages reuse the simulation template by design). All six are already public in the OpenMAIC repository; the slide contract is reproduced here.

\begin{table}[!ht]
  \centering
  \caption{The six system prompts in the training mixture, extracted verbatim from the mix-0812 corpus.}
  \label{tab:sysprompts}
  \small
  \begin{tabular}{@{\hspace{5pt}}lrl@{\hspace{5pt}}}
    \toprule
    Prompt & Length (words) & Used by \\
    \midrule
    Slide content contract & $\sim$310 & all slide requests \\
    Simulation widget template & $\sim$1{,}700 & simulation + learning pages \\
    Interactive diagram template & $\sim$700 & diagrams \\
    Educational game template & $\sim$2{,}200 & games \\
    Code playground template & $\sim$1{,}100 & code playgrounds \\
    3D visualization template & $\sim$2{,}700 & 3D visualizations \\
    \bottomrule
  \end{tabular}
\end{table}

\subsection*{Slide content contract}
\input{apx_sys_slide}


\input{apx_external}

\clearpage
\section{Chinese-Language Production Examples}
\label{app:zhshowcase}
\begin{figure}[!ht]
  \centering
  \includegraphics[width=\linewidth]{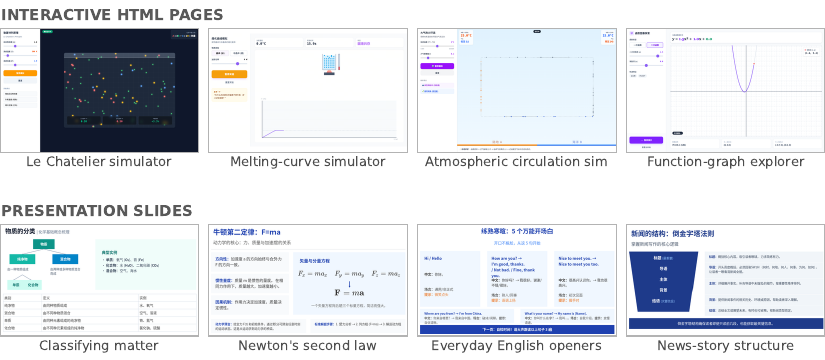}
  \caption{Chinese-language \model output from live production traffic on OpenMAIC, selected with the same criteria as Figure~\ref{fig:showcase}. Top: interactive HTML pages---a Le Chatelier principle particle simulator, a three-step terrain explorer, an atmospheric-circulation simulator, and a function-graph explorer (captured mid-demo). Bottom: slides from generated decks---classifying matter, Newton's second law, everyday English openers, and news-story structure.}
  \label{fig:showcase-zh}
\end{figure}

Figure~\ref{fig:showcase-zh} complements Figure~\ref{fig:showcase} with Chinese-language artifacts sampled from the same live production traffic on OpenMAIC. The mix of modalities and subjects---chemistry, geography, mathematics, physics, English, and writing---mirrors the distribution of learner demand on the platform.

\input{10_contributors}

%% file: 8_future.tex
\section{Discussion and Future Directions}
\label{sec:future}

\model generates learning environments; it does not yet adapt them to the learner. This appendix sketches the direction we are building toward---generation as one stage of a learner-aware pipeline---and lists the problems we consider still open. We present the pipeline as a roadmap rather than a result: none of its stages has been validated end to end.

\paragraph{A four-stage personalization pipeline.}
In the \emph{Evidence} stage, multimodal signals are captured while the student works with the material: eye tracking (fixation durations within areas of interest, regression paths, pupil dilation) for visual attention and cognitive load; wearable sensors (heart-rate variability, electrodermal activity, EEG) for arousal, stress, and fatigue; and software telemetry (dwell times, navigation paths, interaction logs) for explicit behavior. A \emph{Sensemaking} stage translates these low-level signals into a dynamic learner profile of cognitive, affective, and attentional states---structured models for telemetry, sequence models for physiological and gaze time series, or LLMs for cross-modal reasoning. A \emph{Visual Guide} stage then maps the profile to presentation strategies grounded in multimedia-learning theory: high cognitive load triggers lower information density and highlighted anchors, while a conceptual bottleneck triggers stepwise visual decomposition and interactive micro-widgets. These strategies are expressed as brief-level requirements---exactly the input interface \model already consumes---so the final \emph{Generation} stage produces tailored slides and interactive pages without architectural change.

\paragraph{Open problems.}
Four issues from the current system frame our next steps. \textbf{(1) One-big-round at 27B}: the zero-forgetting-tax recipe is validated on 4B only; its 27B confirmation is pending. \textbf{(2) Code rubric strictness}: the hardened reward prices code playgrounds more harshly than its predecessor (53.6 vs.\ 59.1 for the same recipe trained earlier), and it is not yet clear how much of that gap is real quality loss rather than judge calibration. \textbf{(3) Defects the reward does not price}: language mixing and occasional element stacking survive both rounds of manual testing unchanged---consistent with the lesson of the game episode, dimensions absent from the reward do not fix themselves, and extending the reward to typography and language consistency is straightforward in principle. \textbf{(4) Cross-page coherence}: theme drift across the pages of a multi-page course remains; per-page quality does not yet compose into per-course quality, likely requiring course-level context or constraints rather than page-level training alone.

%% file: apx_sys_slide.tex
{\small\begin{verbatim}
You are OpenMAIC slide content generator. Return only one valid JSON object for a
  single 16:9 slide.
Required top-level shape:
  {"elements":[...],"background":{"type":"solid","color":"#ffffff"}}.
For text, shape, image, table, chart, latex, and video elements, include type, id,
  left, top, width, height, and rotate (usually 0). These position fields must be
  JSON numbers without quotes on a 1000x562 canvas; never use percentages, CSS
  layout objects, markdown, or prose.
Line elements use this exact shape: {"type":"line","id":"...","left":0,"top":0,"wi
  dth":2,"start":[0,0],"end":[100,0],"style":"solid","color":"#000000","points":["",
  ""]}. Width is stroke thickness. Do not put height or rotate on line elements. Do
  not make style an object. Do not create shape elements with start/end/points; if
  you need start/end/points, the element type must be line.
Use supported element types only: text, shape, line, image, table, chart, latex,
  video. Do not use rect, circle, group, card, vector_graphic, or layout element
  types. Use shape with path/viewBox/fill for rectangles and circles.
Table elements must use the renderer schema exactly: {"type":"table","id":"...","l
  eft":0,"top":0,"width":400,"height":160,"rotate":0,"outline":{"width":1,"style":"s
  olid","color":"#cbd5e1"},"colWidths":[0.5,0.5],"cellMinHeight":36,"data":[[{"id":"
  cell_1","colspan":1,"rowspan":1,"text":"Header","style":{"bold":true}}]]}. Do not
  use data strings, {header, rows}, column objects, row objects, colSpan, or
  rowSpan.
Chart elements must use chartType one of bar, column, line, pie, ring, area,
  radar, scatter and data exactly as
  {"labels":["A","B"],"legends":["Series"],"series":[[1,2]]}. Do not use donut; use
  chartType "ring" for donut-style charts. Do not use data arrays of
  {label,value,color}, showLegend, showDataLabels, or title fields.
For assigned assets, create image elements with src equal to the asset id such as
  "img_1" and fixedRatio true. Do not redraw assigned images.
\end{verbatim}}

%% file: apx_external.tex
\section{External Flagship Evaluation}
\label{app:external}

This appendix reports the zero-shot (\emph{slim-contract}) slide results for the external flagships referenced in Section~\ref{sec:eval}; their HTML-500 scores and \emph{full-specification} slide results appear in Table~\ref{tab:main}. The two prompt settings differ in a way that matters, so we define them once.

\paragraph{Two prompt settings.}
All slide numbers for external models come through the same harness as our internal runs (same topics, renderer, and judge); what differs is the \emph{system prompt}:
\begin{itemize}[leftmargin=1.5em,itemsep=2pt]
  \item \textbf{Slim contract (the training and serving interface).} The ${\sim}$310-word slide contract of Appendix~\ref{app:sysprompts}---element types, geometry fields, background---with no field-level schema, no examples, no style rules. This is the \emph{only} slide prompt \model ever sees: it is the system prompt of the SFT mixture and of production serving. The schema---required style keys, payload field names, data shapes---is internalized in the weights during training; a serving-side normalization pass (default-filling required style keys, mapping common field aliases) cleans up residual variance for every model alike.
  \item \textbf{Full specification (the distillation-teacher interface).} The 34\,KB design specification our slide data pipeline gives to the teacher model: a complete field table with types and defaults, full few-shot scene-graph examples, style and typography rules, and a pre-render check list. No external flagship has seen this document unless we hand it over explicitly.
\end{itemize}
External flagships therefore run slide-std twice: under the slim contract (zero-shot; Table~\ref{tab:external} below) and under the full specification (Table~\ref{tab:main} in the main text---the strongest fair condition we can hand a model that has not learned the contract). External outputs under both settings pass the same schema normalization as our serving stack; outputs that remain unrenderable score zero. HTML-500 needs no such split---its system prompts (production templates per sub-type) are supplied verbatim to every model.

\begin{table}[!ht]
  \centering
  \caption{External flagships on slide-std under the slim contract (zero-shot): fidelity and layout, each on the judge's anchored 0--5 scale mapped to 0--100 ($\times$20), Avg their mean; outputs that remain unrenderable after production normalization score zero. Reference rows: the released \model models, which train and serve on this 310-word contract. GLM-5.3 and Claude Opus 4.8 were measured under the full specification only (Table~\ref{tab:main}).}
  \label{tab:external}
  \small
  \begin{tabular}{lccc}
    \toprule
    Model & Fid & Lay & Avg \\
    \midrule
    GPT-5.4 & 59.2 & 50.7 & 54.9 \\
    Qwen3.8-Max & 25.3 & 22.5 & 23.9 \\
    DeepSeek-V4-Pro & 50.0 & 45.7 & 47.8 \\
    Gemini 3.6 Flash & 23.2 & 15.3 & 19.2 \\
    \midrule
    \emph{\model-4B (reference)} & \emph{82.8} & \emph{67.3} & \emph{75.1} \\
    \emph{\model-27B (reference)} & \emph{93.0} & \emph{74.3} & \emph{83.7} \\
    \bottomrule
  \end{tabular}
\end{table}

\paragraph{Readings.}
\textbf{(1) The slim condition is the deployment reality for un-trained models.} Zero-shot, every external model names the text payload field \texttt{text} where the schema requires \texttt{content} (1{,}956 of 1{,}958 text elements for GPT-5.4) and none emits the required style keys; normalization restores renderability, but not the conventions that decide quality. The best slim score (GPT-5.4, 54.9) sits sixteen points below our SFT-only 4B (70.8): the contract is a learned data convention, not knowledge a model can deduce---which is what post-training is for.

\textbf{(2) What the full specification buys.} Handed the 34\,KB document (Table~\ref{tab:main}), Gemini 3.6 Flash jumps $+58.5$ points (19.2 $\rightarrow$ 77.7, above \model-4B); GPT-5.4 gains $+15.9$ but its layout stays flat (50.7 $\rightarrow$ 51.7) even as fidelity reaches 90.0---the same content-is-free, composition-is-learned split the bare Qwen3.8-27B base shows (fidelity 91.7, layout 41.5; Section~\ref{sec:task}). Qwen3.8-Max gains the most of all ($+59.8$, 23.9 $\rightarrow$ 83.7): it is the flagship of the family our base belongs to, and with the field table in hand it emits contract-clean scene graphs (5/120 unrenderable, all token-budget truncations). DeepSeek-V4-Pro \emph{loses} ground under the full spec ($-4.5$): the specification invites longer scenes, and 49/120 outputs truncate at the 16{,}384-token budget.

\textbf{(3) The generous condition closes the slide gap for exactly one model.} Qwen3.8-Max ties \model-27B at 83.7---but it is the same-family flagship reading the full 34\,KB specification at inference, whereas \model-27B carries the contract in its weights and needs the 310-word training prompt alone; every other flagship lands at or below Claude Opus 4.8's 79.5. And no model leads both modalities: Qwen3.8-Max's HTML-500 average is 35.3 with 204 of 500 pages dead at the probe; Claude Opus 4.8's best-external 67.2 comes with 19 dead pages and GPT-5.4's 66.0 with 13, while \model-27B holds 63.7 with zero.

%% file: 10_contributors.tex
\section{Contributors}
\label{app:contributors}

\model is a joint effort between CogEvol Inc.\ and Tsinghua University.

\paragraph{Core contributors.}
Shangqing Tu\textsuperscript{$\ast$},
Daniel Zhang-Li\textsuperscript{$\ast$},
Yucheng Wang\textsuperscript{$\ast$},
Shiyu Gan,
Yanpeng Wang,
Huiqiang Rong,
Mofei Chen,
Shen Yang

\paragraph{Advisors.}
Jifan Yu\textsuperscript{$\dagger$},
Juanzi Li,
Bin Xu,
Lei Hou,
Huiqin Liu,
Yu Zhang

\paragraph{Contributors.}
Yini Chen,
Yinuo Duan,
Binglin Liu,
Ye He,
Danqi Zheng,
Zhanxin Hao,
Yuxuan Wu,
Mengting Tao,
Yuqiu Liu

\smallskip
\noindent\textsuperscript{$\ast$}\,Tech leads: Shangqing Tu, Daniel Zhang-Li, Yucheng Wang.\\
\textsuperscript{$\dagger$}\,Corresponding author: \href{mailto:yujifan@mail.tsinghua.edu.cn}{yujifan@mail.tsinghua.edu.cn}